\documentclass[10pt,twocolumn,letterpaper]{article}

\usepackage{iccv}              % To produce the CAMERA-READY version
\usepackage{multirow}
\usepackage{graphicx}
\usepackage{makecell}
\usepackage{xcolor}
\usepackage{array}
\usepackage{booktabs}
\usepackage{caption}
\usepackage{mwe}
\usepackage{color}
\usepackage{subcaption}

\definecolor{iccvblue}{rgb}{0.21,0.49,0.74}
\usepackage[pagebackref,breaklinks,colorlinks,allcolors=iccvblue]{hyperref}

\usepackage[capitalize]{cleveref}
\crefname{section}{Sec.}{Secs.}
\Crefname{section}{Section}{Sections}
\Crefname{table}{Table}{Tables}
\crefname{table}{Tab.}{Tabs.}

\title{ZipIR: Latent Pyramid Diffusion Transformer for \\ High-Resolution Image Restoration}

\author{%
   Yongsheng Yu\textsuperscript{1,2\thanks{Work done during an internship at Adobe.}}, Haitian Zheng\textsuperscript{2},  Zhifei Zhang\textsuperscript{2}, Jianming Zhang\textsuperscript{2}, \\ Yuqian Zhou\textsuperscript{2}, Connelly Barnes\textsuperscript{2}, Yuchen Liu\textsuperscript{2}, Wei Xiong\textsuperscript{2}, Zhe Lin\textsuperscript{2},  Jiebo Luo\textsuperscript{1}\\
    \textsuperscript{1}University of Rochester, \textsuperscript{2}Adobe Research 
 }

\begin{document}
\maketitle

\begin{figure*}[t!]
    \small
    \centering
    \begin{minipage}{0.33\textwidth}
        \centering
        \includegraphics[page=1,height=7cm]{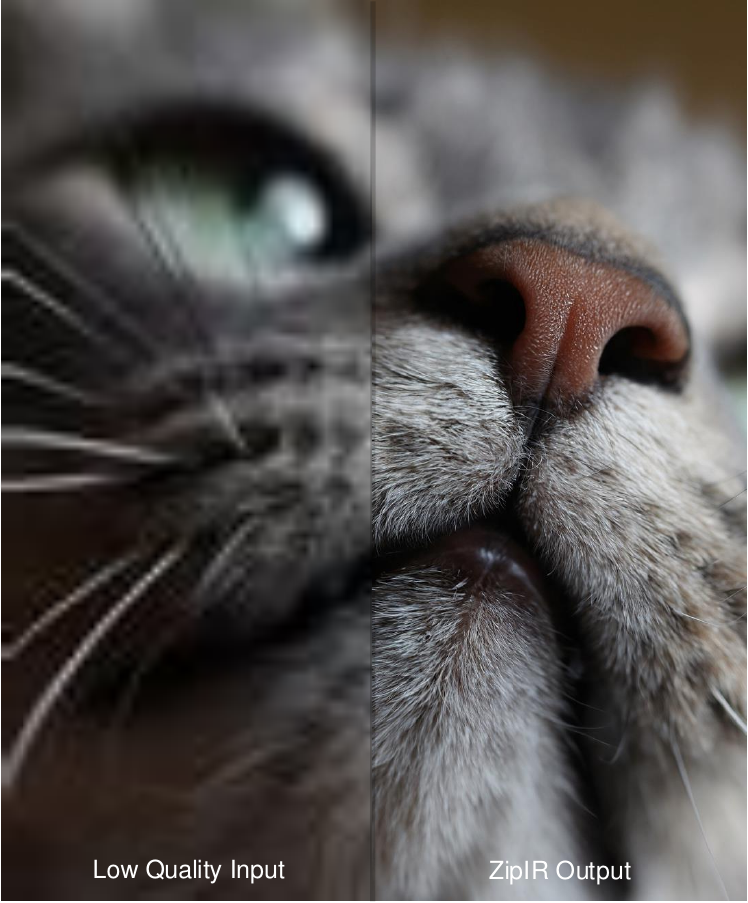} 
        \subcaption{20$\times$ super-resolution at $2048^2$ px resolution.}
    \end{minipage}
    \begin{minipage}{0.33\textwidth}
        \centering
        \includegraphics[page=3,height=7cm]{fig/teaser_figs-crop.pdf} 
        \subcaption{16$\times$ super-resolution at $2048^2$ px resolution.}
    \end{minipage}
    \begin{minipage}{0.33\textwidth}
        \centering
        \includegraphics[page=2,height=7cm]{fig/teaser_figs-crop.pdf} 
        \subcaption{8$\times$ restoration at $2048^2$ px resolution.}
    \end{minipage}
    \vspace{1mm}
    
    \begin{minipage}{0.49\textwidth}
        \centering
        \includegraphics[width=\textwidth]{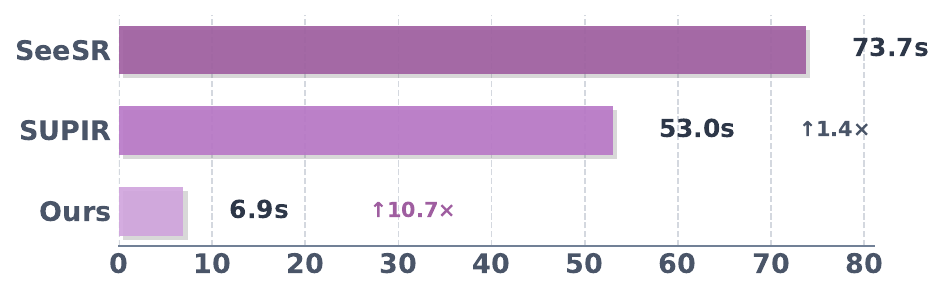} 
        \subcaption{Inference time at $2048^2$ px resolution (in seconds).}
    \end{minipage}
    \begin{minipage}{0.49\textwidth}
        \centering
        \includegraphics[width=\textwidth]{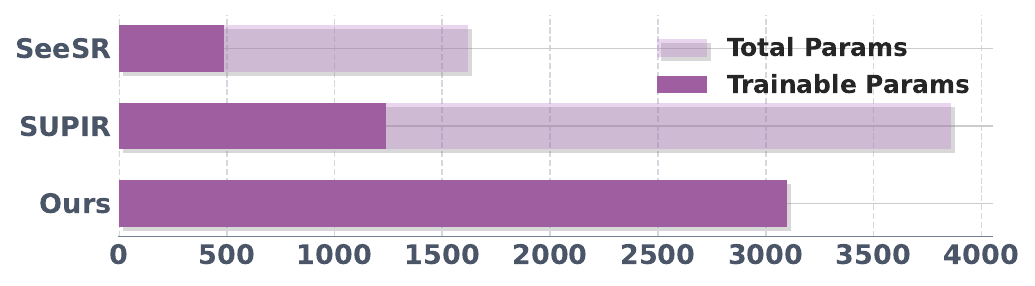} 
        \subcaption{Model scalability measured by the diffusion model parameters (in Millions).}
    \end{minipage}
	% \captionof{figure}{
    % \vspace{-2mm}
    \caption{
    Our ZipIR demonstrates a strong capacity in restoring severely degraded images, such as $20\times$, $16\times$ downsampled or $8\times$ degraded inputs to restore $2048^2$ resolution output. 
    Compared to different diffusion-based methods, ZipIR enjoys (d) an up to 10x running time advantage over SeeSR~\cite{wu2024seesr} while (e) maintaining a higher learning capacity for producing high-quality and ultra high-resolution images from severely degraded inputs.}
    % \vspace{-4mm}
    \label{fig:teaser}
\end{figure*}

%%%%%%%%% ABSTRACT
\begin{abstract}
Recent progress in generative models has significantly improved image restoration capabilities, particularly through powerful diffusion models that offer remarkable recovery of semantic details and local fidelity. 
However, deploying these models at ultra-high resolutions faces a critical trade-off between quality and efficiency due to the computational demands of long-range attention mechanisms.
To address this, we introduce ZipIR, a novel framework that enhances efficiency, scalability, and long-range modeling for high-res image restoration.
ZipIR employs a highly compressed latent representation that compresses image 32×, effectively reducing the number of spatial tokens, and enabling the use of high-capacity models like the Diffusion Transformer (DiT).
Toward this goal, we propose a Latent Pyramid VAE (LP-VAE) design that structures the latent space into sub-bands to ease diffusion training.
Trained on full images up to 2K resolution, ZipIR surpasses existing diffusion-based methods, offering unmatched speed and quality in restoring high-resolution images from severely degraded inputs.
\end{abstract}   

%%%%%%%%% BODY TEXT

\vspace{-4mm}
\section{Introduction}
\label{sec:intro}

Recent advanced generative models, such as GANs~\cite{gan} and diffusion models~\cite{ho2020denoising, rombach2022high}, have dramatically improved image restoration (IR). These models leverage long-range context modeling~\cite{vaswani2017attention, wang2018non, alexey2020image}, enhanced architectural designs~\cite{ho2020denoising, nichol2021improved}, and greater model capacity to effectively restore complex image structures from severely degraded or downsampled inputs. However, existing IR methods, often relying on UNet-based diffusion models~\cite{rombach2022high,podell2023sdxl}, are pretrained on an $8\times$ compressed latent space. While being effective, they face efficiency challenges when restoring ultra high-resolution outputs, due to the quadratic computational demands associated with the number of spatial tokens.

Not only scaling up the resolution of models is challenging, but deploying such models for ultra high-resolution IR also presents significant challenges.
At a high level, there seems to be a fundamental dilemma between quality and efficiency.
On one hand, long-range attention modeling is crucial for both visual understanding~\cite{alexey2020image,radford2021learning,hua2024mmcomposition,hua2024finecaption} and synthesis~\cite{ho2020denoising,cai2023retinexformer,cai2022mask,chen2018pixelsnail}, facilitating the recent success of the Diffusion Transformer (DiT)\cite{peebles2023dit} in both image and video generation\cite{hatamizadeh2025diffit,DBLP:journals/corr/abs-2401-03048}.
On the other hand, such capacity comes with an extensive computational overhead with the number of spatial tokens, dramatically limiting the scalability of these methods for ultra high-resolution IR. 
As shown in~\cref{fig:teaser}, existing diffusion-based IR methods~\cite{wu2024seesr,yu2024supir} take approximately one minute to process a 2K image, and if tiled-based inference is employed, the runtime increases further. This limitation also impedes the exploration of more scalable models like DiT.

In this work, we introduce \textbf{ZipIR}, a novel framework designed to enhance model capacity, efficiency, and long-range modeling for high-quality, high-resolution diffusion-based image restoration.
We start with building a highly compressed latent representation~\cite{rombach2022high,esser2021taming} with a spatial downsampling factor of $f=32$. Differing from existing methods~\cite{lin2024diffbir,wang2024stablesr,wu2024seesr,yu2024supir}, our design effectively reduces the number of latent tokens and offers benefits: it enables the use of advanced models like DiT, facilitates training on the entire images rather than local patches for improved holistic modeling, and increases efficiency during both training and inference phases. As a result, ZipIR achieves up to $10$ times faster inference than SeeSR~\cite{wu2024seesr} at 2K resolution and provides enhanced restoration for severely degraded inputs (downsampled by $20\times$ or $16\times$).

However, designing the $f32$ latent space for image restoration introduces several challenges. 
First, a naively trained latent space is susceptible to minor perturbation and low-level degradation~\cite{szegedy2013intriguing,moosavi2016deepfool}, complicating the restoration process on latent space and leading to instability.
Second, decoding from such a compressed latent code often distorts essential low-level details. To address these issues, we develop a novel Latent Pyramid VAE (LP-VAE) inspired by the Laplacian pyramid representation from image processing literature~\cite{burt1987laplacian}. 
We train the latent space sequentially from lower to higher resolutions: early channels encode lower-resolution information, and subsequent channels capture residual details necessary for reconstructing high-resolution images. This sub-band decomposition effectively separates high-level image structures from low-level details. It ensures that the low-level degradation primarily affects the finer-level latent features, while the coarser-level codes remain consistent, thereby simplifying the learning process for the diffusion model.

Building upon LP-VAE, we finally designed a novel architecture based on DiT~\cite{peebles2023dit} to scale the capacity of the diffusion model for high-resolution IR. Trained on the entire image at up to 2K resolution and benefiting from the long-range modeling capacity, our method shows a stronger generation capacity, capable of upsampling across a wide range of scale factors (from 8$\times$ to 20$\times$) directly at 2K resolution while achieving a significant speedup over previous diffusion-based image restoration methods including the recent SUPIR~\cite{yu2024supir}, without sacrificing the quality.

To summarize, we introduce ZipIR, a novel diffusion-based framework designed for high-quality and efficient high-resolution image restoration. 
Leveraging the highly compact and structured LP-VAE latent space, along with a scaled-up diffusion model trained on full high-resolution images, ZipIR seamlessly reconstructs 2K images with both globally coherent structures and fine local fidelity from heavily degraded inputs, outperforming existing diffusion-based approaches in both efficiency and quality.

\begin{figure*}[t]
    \centering
    \includegraphics[width=\linewidth]{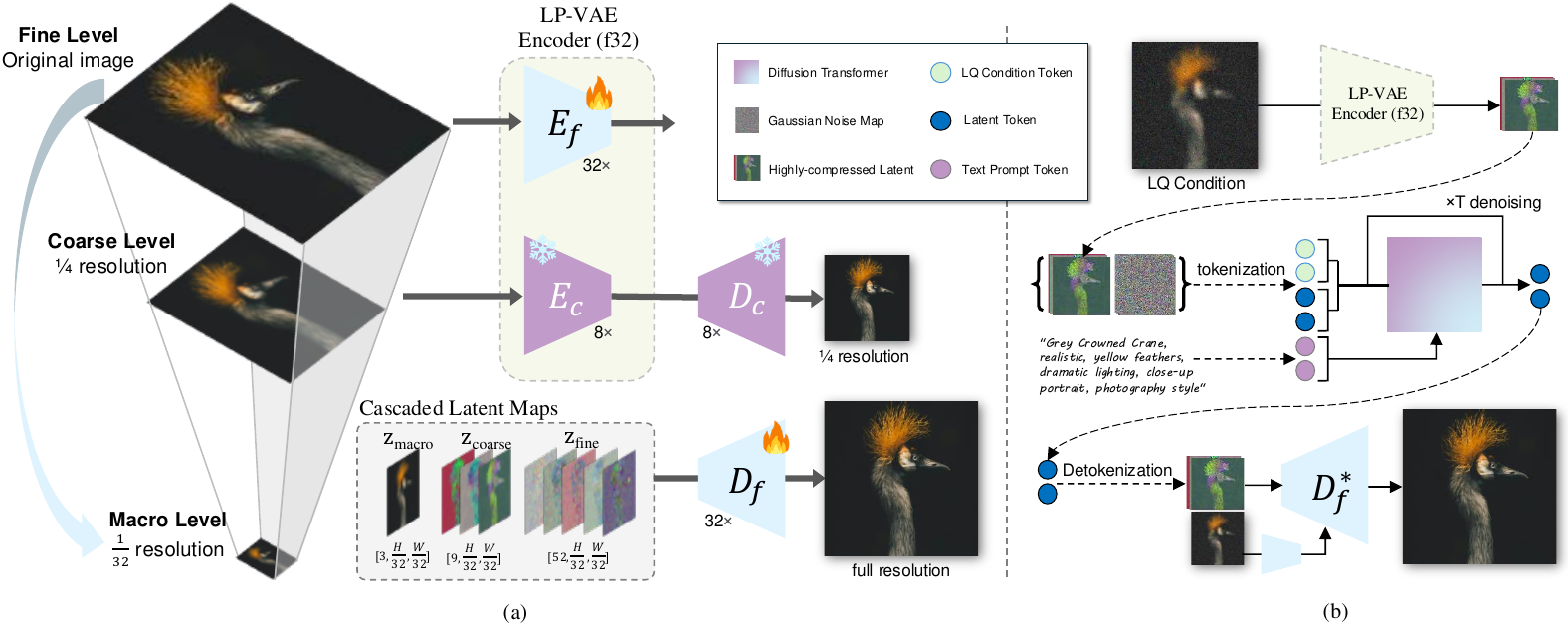}
    % \vspace{-5mm}
    % \small
    \caption{\textbf{Overview of ZipIR}: (a) Latent Pyramid VAE (LP-VAE) compresses raw images into a 32$\times$ downsampled latent space through a pyramidal design that captures sub-band information across multiple scales (32$\times$, 4$\times$, original), ensuring a high compression rate while maintaining a well-structured latent space with improved representation invariance under degradation. (b) Our transformer diffusion model is trained and operates on the compressed latent space of the entire image, supporting resolutions up to $2048^2$ pixels, enriching semantic understanding and synthesis of holistic structure. Furthermore, the pixel-space decoder $D_f^\ast$ (\cref{sec:method:lpvae}) further enhances restoration quality.
    }
    % \vspace{-4mm}
    \label{fig:main-pipeline}
\end{figure*}

\section{Related Work}
\subsection{High-Resolution Image Restoration}

High-resolution image restoration (HR-IR) aims to enhance degraded images, often requiring models capable of generating fine-grained details at high fidelity. Early approaches to image restoration targeted specific degradations independently, such as super-resolution (SR)~\cite{dong2015image,yang2008image}, denoising~\cite{zhao2022manet,chen2016trainable}, and deblurring~\cite{xu2014deep,xu2014inverse}, often relying on fixed assumptions about degradation patterns. While effective within constrained conditions, these methods lacked the flexibility to handle real-world complexities. Recently, blind IR methods have gained popularity~\cite{wang2021realesrgan,zhang2021bsrgan,yu2024supir}, integrating multiple restoration tasks within unified frameworks that can generalize across diverse degradations, as exemplified by DiffBIR~\cite{lin2024diffbir}.

GAN-based methods have achieved realistic restoration results on real-world degraded images~\cite{zhang2021bsrgan,wang2021realesrgan}. However, these methods have limitations, particularly in preserving fine details under extreme scaling factors. Diffusion-based approaches like StableSR~\cite{wang2024stablesr} and SUPIR~\cite{yu2024supir}, which leverage pre-trained models like Stable Diffusion~\cite{rombach2022high,podell2023sdxl}, have demonstrated notable improvements in restoration quality through multi-step processes, though these can be computationally intensive.

Scaling up restoration models has shown promise, especially with advancements in large-scale architectures~\cite{yu2024supir}. Our proposed method leverages the scalability of diffusion transformers~\cite{peebles2023dit} to tackle the complex, high-dimensional nature of HR-IR.

\subsection{Efficient Diffusion Models}

Diffusion models are powerful generative tools but face challenges with high computational demands and slow sampling speeds, limiting their practicality~\cite{ho2020denoising, rombach2022high}. Sampling-efficient methods~\cite{song2020denoising, yin2024improved, song2023consistency, liu2023instaflow} reduce the number of sampling steps, thereby shortening runtime, while model-based optimizations refine model architecture, using strategies like pruning~\cite{DBLP:conf/nips/FangMW23, castells2024ld} and linear-complexity modules~\cite{liu2024linfusion, fei2024diffusion} to create faster, more compact models. As diffusion models scale for high-resolution tasks, memory limitations and inference latency also become pressing issues. Our method addresses these with LP-VAE, a compact latent encoding approach that intensifies compression, reducing the spatial dimensions of feature maps and thus easing the computational load for high-resolution image restoration.

\vspace{-2mm}
\section{Methodology}
\subsection{Latent Pyramid VAE (LP-VAE)}
\label{sec:method:lpvae}

To enable billion-scale DiT models to operate at 2K resolution and beyond, our priority is to optimize latent channel capacity and deepen the latent space mapping by adding more downsampling layers.
This reduces the token count, lowering the quadratic complexity of DiT built on self-attention. 
As spatial compression increases, the spatial resolution of the latent representation shrinks, necessitating a corresponding increase in the latent channel count $C$ to mitigate information loss. For an input image $I \in \mathbb{R}^{3 \times H \times W}$, the encoder maps it to a latent code $\mathbb{Z} \in \mathbb{R}^{C \times \frac{H}{f} \times \frac{W}{f}}$. Despite the increased latent channels, raising the compression ratio still significantly impacts reconstruction quality, as evident from our ablation tests in Table~\ref{tab:ablation-study}.

\noindent\textbf{Pyramid Cascade Encoders}.
Cascading networks have proven effective in other generative models~\cite{pernias2023wurstchen,ho2022cascaded,teng2023relay}, which allows different networks to independently learn representations at different resolutions, optimizing overall pipeline performance. Accordingly, our architecture employs a three-level pyramid VAE encoder to capture fine-level which encodes image high-frequency details, coarse-level features which encode lower-res structures, and macro-level semantics, with cascaded latent codes serving as a highly compressed image representation. The pyramid latent structure is shown on the left side of~\cref{fig:main-pipeline}. 

The fine and coarse-level encoders independently encode representations from different resolutions. For $f=32$, the fine-level encoder operates on the original image $I$ without downsampling, producing a 52-channel latent encoding $z_{\textrm{fine}} \in \mathbb{R}^{52 \times \frac{H}{32} \times \frac{W}{32}}$. The coarse-level encoder captures lower-resolution features with an 4$\times$ downsampled input, $I_{\downarrow 4} \in \mathbb{R}^{3 \times \frac{H}{4} \times \frac{W}{4}}$, resulting in 9-channel latent encoding $z_{\textrm{coarse}} \in \mathbb{R}^{9 \times \frac{H}{32} \times \frac{W}{32}}$. 
To incorporate macro-level semantics, we use downsampled $I_{\downarrow 32}$ as a 3-channel image $z_{\textrm{macro}} = \frac{I_{\downarrow 32} - \mu}{\sigma}$, where $\mu$ and $\sigma$ are the mean and standard deviation calculated from the entire training dataset. Finally, the concatenated latent code across all levels, denoted by $\mathbf{z} = [z_{\textrm{macro}}; z_{\textrm{coarse}}; z_{\textrm{fine}}]$, serves as the final highly-compressed 64-channel representation.

\begin{figure*}[t]
    \small
    \centering
    \includegraphics[page=1, width=0.96\linewidth]{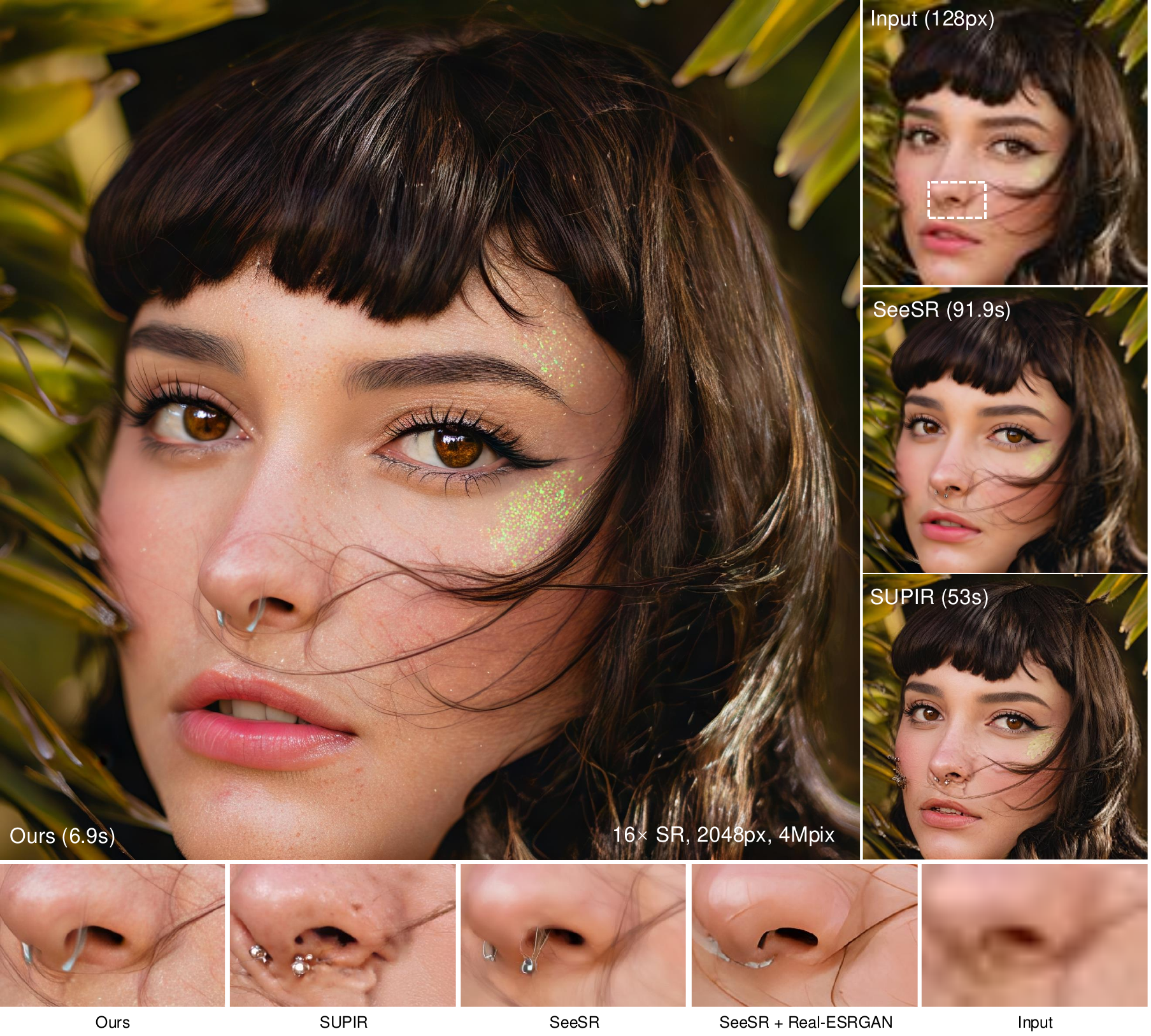}
    \vspace{-2mm}
    \caption{Our DiT-based ZipIR serves as a 16$\times$ upsampler, completing super-resolution from 128$^2$ to 2048$^2$ in only $6.9$ seconds. We compare it with SUPIR~\cite{yu2024supir} and SeeSR~\cite{wu2024seesr}, evaluating both their direct 2K inference and SeeSR's default 512px output upscaled to 2K via Real-ESRGAN~\cite{wang2021realesrgan}. The gold dust on the face is real, not an artifact.}
    \label{fig:qualitative_comparison_lady}
     \vspace{-3mm}
\end{figure*}

\begin{figure*}[t]
    \small
    \centering
    \includegraphics[page=6,width=0.96\linewidth]{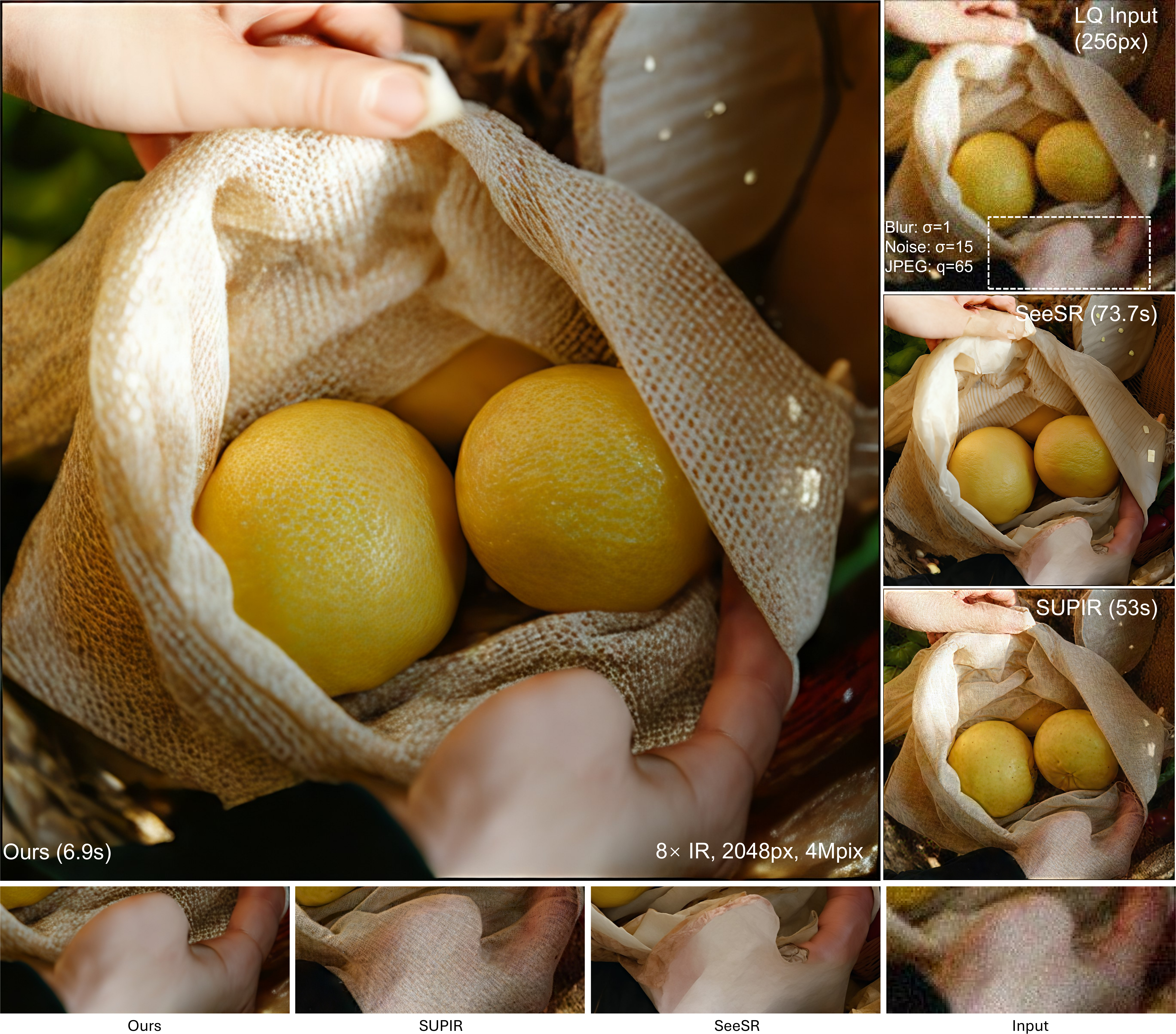}
    \caption{Our DiT-based ZipIR functions as an 8× upsampler, enhancing images from 256$^2$ to 2048$^2$ in just 6.9 seconds, while simultaneously restoring details through deblurring, denoising, and JPEG artifact removal. The input image is degraded with Gaussian blur ($\sigma = 1$), noise ($\sigma = 15$), and JPEG compression ($q = 65$). The reconstructed hand retains biological features without merging with the sack texture.}
    \label{fig:qualitative_comparison_scene}
\end{figure*}

\noindent\textbf{Progressive Training}.
We employ a progressive training approach. 
Training begins with the coarse-level encoder $E_c$, which requires a decoder $D_c$ to reconstruct the 12-channel latent $[z_{\textrm{macro}};z_{\textrm{coarse}}]$ into pixel space. 
After completing this training stage, the coarse-level decoder $D_c$ is discarded. Progressively, the next stage involves training the fine-level autoencoder to achieve full-level compression. 
The left side of~\cref{fig:main-pipeline} illustrates this stage, the fine-level decoder $D_f$ is trained to reconstruct from a 64-channel latent $\mathbf{z}$ back to pixel space, while the coarse-level encoder remains frozen.

For both training phases, we use a combination of discriminator loss and LPIPS loss as recommended in~\cite{esser2021taming}. 
Based on empirical findings, we observed that attention layers did not significantly improve performance and added unnecessary overhead for both encoding and decoding. Therefore, our entire LP-VAE is designed as a pure convolutional network. 
Once training is complete, $E_c$ and $E_f$ serve as sub-networks of the LP-VAE Encoder, cascading three types of compressed mappings into a highly-compression representation $\mathbf{z} \in \mathbb{R}^{64 \times \frac{H}{32} \times \frac{W}{32}}$. Finally, a non-pyramidal decoder network $D_f$ decodes $\mathbf{z}$ to obtain the RGB image. 

\noindent\textbf{Pixel-aware Decoder-only Finetuning}.
Reconstructing from our highly compressed latents space achieves notable quality in high-resolution image reconstruction. However, without access to the full-resolution input, the decoder remains suboptimal for image processing applications, particularly restoration tasks that demand high pixel fidelity and detailed quality. Therefore, after obtaining the LP-VAE with its encoder $E_f$, decoder $D_f$, and the associated 64-channel latent space, we incorporate pixel-level details through skip connections to add spatial information during LP-VAE decoding, leading to a pixel-aware decoder $D_f^*$.

To capture spatial features, we replicate an LP-VAE sub-encoder, $E_{f}$, initializing it with weights from the pretrained $E_{f}$. This degradation-aware feature extractor specifically handles degraded images, such as those blurred, noisy, or affected by JPEG artifacts. Additional residual layers are inserted between upsampling blocks in each layer of the LP-VAE decoder $D_f$ to pass multiscale spatial information from the degradation-aware feature extractor, effectively capturing details from low-quality inputs. To train the pixel-aware decoder $D_f^*$, we freeze the original $E_f$, but unlock the degradation-aware feature extractor and the entire decoder. With image reconstruction as the learning objective, this set-up enables the decoder to learn how to utilize low-quality images to complement the highly compressed latent code with pixel-level details. Note that decoder fine-tuning can occur after diffusion training to enhance quality, since the latent space is not altered by freezing $E_f$. 

\subsection{Diffusion Transformer for Image Restoration}

With the LP-VAE trained, we use its encoder to represent an input image \( I \). Our model, a scaled-up diffusion transformer architecture of 3B parameters, $G$, is optimized for high-resolution image restoration.
% \subsubsection{Conditioned Synthesis}
As illustrated in ~\cref{fig:main-pipeline}b, our framework uses two conditioning inputs: a low-quality image \( I_{\textrm{LQ}} \in \mathbb{R}^{3 \times H \times W} \) and a text embedding \( y \), integrating both visual and semantic guidance for restoration.

\noindent \textbf{Low-Quality Image Conditioning. }
Unlike traditional restoration methods relying on pixel-level low-quality (LQ) inputs, we resize \( I_{LQ} \) to the target resolution, compress it using our LP-VAE Encoder, and concatenate it with a noisy latent \( z_t \). Despite sharing the same latent space, these latents differ significantly, necessitating separate, parameter-independent Patch Embedders for tokenization, which are arranged in parallel within the token sequence.

\noindent \textbf{Text Semantic Guidance. }
Text embeddings aid in reconstructing degraded images by refining regions based on contextual cues~\cite{yu2024supir}. We train $G$ on paired text-image data, where the text prompt is a caption of the original image, encoded by T5 language model~\cite{raffel2020exploring}, and integrated via cross-attention layers within the Diffusion Transformer. To support classifier-free guidance, we randomly drop the text embedding with a probability of $0.05$ during training. Additionally, we annotated low-quality images with negative prompts, enabling the model to produce clearer, more realistic outputs during inference with added negative text prompts. 
The effects of varying text prompt strengths are further analyzed in our supplementary material.

\noindent\textbf{Learning HR-IR with DiT}.
We train our model on synthetic data degraded using methods similar to~\cite{wang2021realesrgan}, resizing the generated low-quality images to match the high-quality images for training. This approach aligns with our focus on high-resolution restoration while ensuring robustness to various degradation patterns.
% \noindent \textbf{High-resolution Patch Training. }
Leveraging the highly-compressed latent space, we only need to process a 64$\times$64 latent map even for an original image resolution of $2048$. This allows us to train DiT models on \textbf{high-resolution images} more effectively than existing methods~\cite{yu2024supir,wu2024seesr}, facilitating the global-range semantic understand. During DiT training, we mix crop patches ranging from $512^2$ to $4096^2$.

\begin{table}[t]
\centering
\small
\caption{Quantitative comparison of image restoration methods under various degradation types. ``Mixture degradation" denotes that the input image undergoes 8$\times$ downsampling, Gaussian blur with $\sigma=2$, noise with $\sigma=40$, and JPEG artifacts with $p=50$.}
 \vspace{-3mm}
\label{tab:main_quantitative}
\resizebox{\linewidth}{!}{%
\begin{tabular}{llccccc}
\toprule
 & Method & PSNR$\uparrow$ & LPIPS$\downarrow$ & FID $\downarrow$ & pFID $\downarrow$ & KID$_{\times 10^3}$ $\downarrow$ \\ \midrule
\multirow{7}{*}{\rotatebox{90}{SR (16$\times$)}} 
    & Real-ESRGAN~\cite{wang2021realesrgan} & 25.55 & 0.5535 & 19.32 & 29.12 & 3.23 \\
    & StableSR~\cite{wang2024stablesr}      & \textbf{26.41} & 0.5683 & 24.40 & 54.22 & 8.47 \\
    & DiffBIR~\cite{lin2024diffbir}         & 25.92 & 0.4405 & 21.13 & 29.90 & 4.40 \\
    & SeeSR~\cite{wu2024seesr}              & 25.22 & 0.4321 & 13.20 & 21.20 & 1.23 \\
    & SUPIR~\cite{yu2024supir}              & 23.85 & 0.4377 & 15.23 & 20.55 & 0.81 \\
    & Ours                                  & 24.44 & \textbf{0.3978} & \textbf{9.89} & \textbf{18.17} & \textbf{0.63} \\ \midrule
\multirow{7}{*}{\rotatebox{90}{SR (8$\times$)}} 
    & Real-ESRGAN~\cite{wang2021realesrgan} & 27.47 & 0.4122 & 10.52 & 21.90 & 1.96 \\
    & StableSR~\cite{wang2024stablesr}      & \textbf{28.93} & 0.4238 & 5.17 & 19.51 & 0.29 \\
    & DiffBIR~\cite{lin2024diffbir}         & 28.03 & 0.3503 & 9.26 & 18.03 & 1.93 \\
    & SeeSR~\cite{wu2024seesr}              & 27.77 & 0.3444 & 4.35 & 17.05 & 0.43 \\
    & SUPIR~\cite{yu2024supir}              & 26.35 & 0.3508 & 7.25 & 15.74 & 0.80 \\
    & Ours                                  & 27.86 & \textbf{0.3374} & \textbf{3.24} & \textbf{13.95} & \textbf{0.02} \\ \midrule
\multirow{6}{*}{\makecell{\rotatebox{90}{Mix. Degradation}}}
    & Real-ESRGAN~\cite{wang2021realesrgan} & 22.24 & 0.5919 & 73.32 & 76.08 & 36.07 \\
    & StableSR~\cite{wang2024stablesr}      & 22.15 & 0.7593 & 123.87 & 172.62 & 73.25 \\
    & DiffBIR~\cite{lin2024diffbir}         & \textbf{22.45} & 0.5806 & 59.29 & 64.35 & 26.19 \\
    & SeeSR~\cite{wu2024seesr}              & 22.06 & 0.6085 & 78.09 & 49.72 & 29.47 \\
    & SUPIR~\cite{yu2024supir}              & 21.65 & 0.6335 & 81.14 & 70.35 & 37.75 \\
    & Ours                                  & 20.41 & \textbf{0.5791} & \textbf{35.10} & \textbf{31.08} & \textbf{11.23} \\ \bottomrule
\end{tabular}%
}
 \vspace{-3mm}
\end{table}

\section{Experiments}
For training both LP-VAE and DiT, we use 300 million curated stock images paired with text as the data source. For the coarse-scale sub-model of LP-VAE, we use a batch size of $512$ and run for $50$K iterations. For the fine-scale sub-model, we start with a batch size of $160$ on $512^2$ cropped patches for $100$K iterations, followed by a 1K patch adaptation with a batch size of $32$ on $1024^2$ cropped patches for $50$K iterations. For DiT, we mix resolutions and aspect ratios to sample training images, similar to~\cite{podell2023sdxl}, using a batch size of $128$ over $250$K iterations, with the standard learning objective~\cite{peebles2023dit} guiding the training process. Inference for ZipIR employs the DDIM sampler with 25 denoising steps.

\subsection{Experimental Settings}

In recent benchmarking of IR, medium-resolution samples at 1024$^2$ or 512$^2$ serve as HQ images, often from limited sets with fewer than $250$ images, such as RealSet65~\cite{yue2024resshift} and DrealSR~\cite{wei2020drealsr}. These datasets are not ideally suited for distribution-based evaluation metrics like Fréchet Inception Distance (FID)~\cite{heusel2017gans} due to their scale. To enable more robust benchmarking for IR at higher resolutions, we collected a comprehensive set of $3000$ 2K-resolution photos from Pexels~\cite{Pexels}, facilitating a thorough evaluation across tasks like mixture degradation restoration and super-resolution across varying scale factors from 8$\times$ to 16$\times$. In the Appendix, we present an analysis of the test set.

Quantitatively, we primarily use FID and Kernel Inception Distance (KID)~\cite{binkowski2018demystifying} to measure output distribution realism. Given the HQ resolution of $2048$, we additionally report Patch FID, inspired by~\cite{chai2022any,ren2024ultrapixel}. Text prompting for our model is provided via InternVL-26B~\cite{chen2024internvl}, which generates consistent image caption. We continue to report PSNR, LPIPS~\cite{zhang2018unreasonable} for benchmark purposes, despite the acknowledged misalignment of pixel-wise similarity metrics with human perception in evaluation~\cite{yu2024supir,suvorov2022resolution}. 
No-reference image quality metrics, such as MANIQA~\cite{yang2022maniqa}, are omitted from the main quantitative experiments in Table~\ref{tab:main_quantitative} because they downsample images to 224$^2$, which may not adequately capture high-resolution restoration performance. To provide a more comprehensive evaluation, we employ the real-world LQ dataset RealPhoto60~\cite{yu2024supir}. For a fair comparison—mirroring SUPIR~\cite{yu2024supir}, which downsamples its 1K results to 512$^2$—we downsample our 1K results to 512$^2$ (note that despite differences in model output resolutions, all methods use the same input images). This approach allows us to compare no-reference image quality metrics, including MANIQA~\cite{yang2022maniqa}, CLIP-IQA~\cite{clipiqa}, and MUSIQA~\cite{musiq}.

We conduct three main experiments to demonstrate our method's performance and assess the contributions of each component. First, we benchmark traditional restoration on RealPhoto60~\cite{yu2024supir} and high-resolution image restoration on our newly proposed validation set of $3000$ images (Section~\ref{sec:main_comparison}). 
Second, we analyze the inference efficiency and model parameters among a series of diffusion-based image restoration methods (Section~\ref{sec:efficiency_analysis}).
Third, an ablation study illustrates the effectiveness of each technical component by adding them incrementally (Section~\ref{sec:ablation_study}).

\subsection{Comparison with Existing Methods}\label{sec:main_comparison}

\noindent\textbf{Quantitative Evaluations}. Table~\ref{tab:main_quantitative} illustrates the comparisons analysis of high-resolution ($2048^2$) image restoration across 16$\times$ to 8$\times$ super-resolution scale factor and a kind of mixture degradation by 8$\times$ downsampling, Gaussian blur $\sigma=2$, noise $\sigma=40$, JPEG artifacts $p=50$.
We evaluate the proposed ZipIR and recent advanced image restoration methods via PSNR, LPIPS~\cite{zhang2018unreasonable}, FID~\cite{heusel2017gans}, Patch FID (pFID) and KID~\cite{binkowski2018demystifying}.
The method ZipIR demonstrates strong performance in high-resolution 16$\times$ and 8$\times$ scenarios. For the 16$\times$ super-resolution, ZipIR achieves notable LPIPS and FID improvements ($0.3978$ and $9.89$, respectively), indicating superior perceptual quality and fidelity, while maintaining a competitive PSNR score. Its KID score ($0.63 \times 10^3$) also emphasizes the reduced distributional discrepancy compared to other models. 
For the 8$\times$ super-resolution task, ZipIR continues to show robustness, with the lowest FID ($3.24$) and best LPIPS ($0.3374$), affirming its quality consistency across different scales. 
Under mixed degradation, although the evaluation on pixel-wise similarity of ZipIR is lower, its LPIPS ($0.5791$), FID ($35.10$), Patch FID ($31.08$) and KID ($11.23 \times 10^3$) reflect an ability to preserve perceptual quality and distributional consistency in challenging conditions. 

Furthermore, in no-reference image quality assessment, as in Table~\ref{tab:realphoto60}, our method achieves the best or second-best performance across all metrics. Specifically, while CLIP-IQA~\cite{clipiqa} and MUSIQ~\cite{musiq} scores are slightly lower than those of SUPIR~\cite{yu2024supir}, they remain highly comparable ($0.8154$ vs. $0.8232$ for CLIP-IQA~\cite{clipiqa} and $72.75$ vs. $73.00$ for MUSIQ~\cite{musiq}). Moreover, our method outperforms all others in MANIQA~\cite{yang2022maniqa}, highlighting the effectiveness of our approach in real-world LQ image restoration.

\begin{table}[t]
    \centering
    \renewcommand{\arraystretch}{1.2}
    \caption{Quantitative evaluation of real-world LQ images from RealPhoto60~\cite{yu2024supir} using no-reference image quality metrics.}\label{tab:realphoto60}
    \resizebox{\linewidth}{!}{
    \begin{tabular}{lccccccc}
        \toprule
        Metrics & BSRGAN & Real-ESRGAN & StableSR & DiffBIR & SeeSR & SUPIR & Ours \\
        \midrule
        CLIP-IQA & 0.4119 & 0.5174 & 0.7654 & 0.6983 & 0.7721 & \textbf{0.8232} & \underline{0.8154} \\
        MUSIQ & 55.64 & 59.42 & 70.70 & 69.69 & 72.21 & \textbf{73.00} & \underline{72.75} \\
        MANIQA & 0.1585 & 0.2262 & 0.3035 & 0.2619 & \underline{0.5596} & 0.4295 & \textbf{0.6681} \\
        \bottomrule
    \end{tabular}}
\end{table}

\begin{table}[t]
% \small
\centering
\caption{Efficiency comparison of recent diffusion-based image restoration methods at $2048^2$ resolution, including Neural Function Evaluations (NFEs), latency per denoising step, total inference time per image, and trainable parameters in diffusion models.}
 % \vspace{-3mm}
\label{tab:efficiency}
\resizebox{\linewidth}{!}{%
\begin{tabular}{lccccc}
\toprule
Model & Type & NFEs & \makecell[c]{Denoising\\Latency} (ms) & Inf. Time & \# \makecell[c]{Trainable \\ Param.} \\
\midrule
SeeSR~\cite{wu2024seesr} & UNet & 50 & 1420 & 73.736s & 0.5 B\\
SUPIR~\cite{yu2024supir} & UNet & 50 & 901 & 52.994s & 1.2 B \\ \midrule
Ours & DiT & \textbf{25} & \textbf{250} & \textbf{6.923s} & \textbf{3.1 B}\\
\bottomrule
\end{tabular}}
 % \vspace{-3mm}
\end{table}

\begin{figure*}[htbp]
    \small
    \begin{minipage}[t]{0.49\linewidth}
        \centering
        \includegraphics[page=1, height=0.115\textheight]{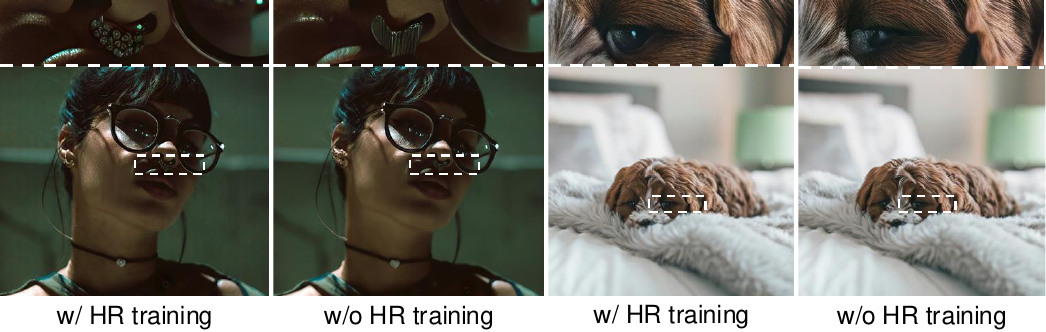}
        % \vspace{-5mm}
        \caption{Effect of HR training for high-res restoration.}
        \label{fig:ablation-qualitative-hr-patch}
    \end{minipage}%
    \hfill
    \begin{minipage}[t]{0.49\linewidth}
        \centering
        \includegraphics[page=2, height=0.115\textheight]{fig/ablation-qualitative-analysis-crop.pdf}
        % \vspace{-5mm}
        \caption{Comparison for w/ and w/o the pixel-aware decoder.}
        \label{fig:ablation-qualitative-pixel-aware-decoder}
    \end{minipage}
    \vspace{-4mm}
\end{figure*}

\noindent\textbf{Qualitative Evaluations}. Figures~\ref{fig:qualitative_comparison_lady} and~\ref{fig:qualitative_comparison_scene} present the visual comparison across existing the most advanced image restoration baselines. For facial portrait restoration at 16$\times$ SR, ZipIR produces sharper, more natural results than competing models, capturing intricate local details like the nose structure and piercings.
In comparison, SUPIR~\cite{yu2024supir} and SeeSR~\cite{wu2024seesr} fall short of ZipIR in preserving clarity, exhibiting noticeable distortions. We also evaluate an alternative approach where SeeSR processes its default resolution (512px), followed by upsampling with the efficient Real-ESRGAN~\cite{wang2021realesrgan}. While this results in slightly sharper outputs compared to SeeSR’s direct 2K inference, it still introduces artifacts, compromising overall visual fidelity.
In the 8$\times$ IR task, as shown in ~\cref{fig:qualitative_comparison_scene}, the LQ input suffers from blur $\sigma=1$, noise $\sigma=15$, and JPEG artifacts $q=65$. SeeSR~\cite{wu2024seesr} introduces over-sharpening artifacts, while SUPIR~\cite{yu2024supir} over-smooths textures, leading to unnatural hallucinations. In contrast, our ZipIR effectively restores fine-grained structures such as skin details, fabric texture, and citrus surfaces while minimizing artifacts.

\begin{table}[b]
\centering
\small
\vspace{-2mm}
\caption{Ablation on our model design, including latent space choices, model scaling, and different diffusion training schemes. For various VAEs, $f$ represents the compression factor, while $c$ denotes the dimensionality of the latent channels.}
 \vspace{-2mm}
\label{tab:ablation-study}
\renewcommand{\arraystretch}{1.2}
\resizebox{0.95\linewidth}{!}{
\begin{tabular}{l c c c}
\toprule
Model & FID $\downarrow$ & pFID $\downarrow$ & PSNR $\uparrow$ \\
\midrule
0.68B DiT & & & \\ 
\quad + f8c4-SDVAE & 30.74 & 53.45 & 28.41 \\ 
\quad + f32c64-SDVAE & 35.83 & 59.47 & \textbf{29.06} \\  
\quad + f32c64-LP-VAE & \textbf{28.14} & \textbf{51.73} & 28.50 \\ \midrule
3B DiT \\
\quad f32c64-LP-VAE & & & \\
\qquad + Pyramid Cascade Encoders & 22.84 & 41.12 & 26.90 \\
\qquad + 1K Patch Adapation &  21.09 & 39.94 & 26.75\\
\qquad + Pixel-aware Decoder & 20.95 & 38.73 & 27.94\\
\quad + HR Training & \textbf{18.05} & \textbf{34.85} & \textbf{27.75} \\
\bottomrule
\end{tabular}}
% \vspace{-3mm}
\end{table}

\subsection{Efficiency Analysis}\label{sec:efficiency_analysis}
Table~\ref{tab:efficiency} summarizes a comparison of ZipIR with several advanced baseline methods in terms of denoising latency, processing time per image, and trainable diffusion model parameters.
ZipIR achieves a much lower denoising latency of $250$ ms, outperforming all baselines, with the closest competitor, SUPIR, showing a latency of $901$ ms.
This efficiency is due to our proposed LP-VAE, which achieves a 32$\times$ compression rate, significantly reducing the input token count in the diffusion transformer.  Even at 2K resolution, each diffusion denoising step requires substantially less time. Consequently, ZipIR demonstrates exceptional efficiency in image processing, taking only $6.92$ seconds per image--a significant improvement over other models like SeeSR ($73.73$ seconds) and SUPIR ($52.99$ seconds).

Our LP-VAE introduces minimal overhead when encoding or decoding 2K images, highlighting its design efficiency. Despite ZipIR's larger model size of $3.1$ billion parameters, it performs inference $10.7$ times faster than SeeSR, which has $1.4$ billion parameters in the diffusion model. These results emphasize the superior efficiency and scalability of ZipIR for practical applications.

\subsection{Effectiveness of Proposed Components}\label{sec:ablation_study}

We quantitatively demonstrate the impact of our proposed components through an ablation study in Table~\ref{tab:ablation-study}. To facilitate the experiments, we sampled $100$ images from a benchmark dataset of $3000$ for the ablation study, performing 8× super-resolution from LQ $128^2$ px to HQ $1024^2$ resolution. We report the metrics FID, Patch FID (pFID), and pixel-wise similarity PSNR. Starting with a baseline $0.68$B DiT model paired with the original f8c4-SDVAE, we observe that switching directly to f32c64-SDVAE results in a decline in FID and pFID. This indicates that naively stacking networks or increasing channel dimensions in VAE does not guarantee robust improvement,
as the latent code is susceptible to low-level perturbation and complicates the diffusion training.
Next, by introducing our f32c64-LP-VAE, we achieve notable performance gains across all metrics, underscoring the impact of an optimized VAE design. Due to the lack of a pre-trained f32c64-SDVAE checkpoint, we trained it with the same settings as f32c64-LP-VAE for a fair comparison.
Scaling up to a $3$B DiT model, we incrementally add each of our proposed components. Notably, as we scale up the diffusion transformer, we observe significant boosts in perceptual quality and fidelity. Each addition, from Pyramid Cascade Encoders progressively enhances performance, with consistent reductions in FID and pFID alongside increases in PSNR.

\noindent\textbf{HR Training}.
Our high-compression LP-VAE encoder (f32) allows DiT models to be trained on global image above 2K resolution. We conduct a qualitative study to demonstrate the effect of the HR training technique. As illustrated in \cref{fig:ablation-qualitative-hr-patch}, HR training facilitates sharper and more accurate local details, such as the structure of accessories and textures of fur, compared to its counterpart.  

\noindent\textbf{Pixel-aware Decoder}.
The pixel-aware decoder is introduced as a complementary module to restore the spatial information of the input image at the pixel level. This proposed module enables ZipIR to capture spatial details directly from the original image, rather than relying solely on latent-level information. As shown in~\cref{fig:ablation-qualitative-pixel-aware-decoder}, the use of the pixel-aware decoder enhances clarity in textual and structural details, demonstrating its effectiveness.

\vspace{-2mm}
\section{Conclusion and Future Work}
\label{sec:conclusion}
\vspace{-2mm}
We present ZipIR, a framework that tackles efficiency, scalability, and quality in ultra-high-resolution image restoration. We developed the Latent Pyramid VAE (LP-VAE) to compress images into a structured latent space, enabling the training of the high-capacity Diffusion Transformer (DiT) on entire images. Tested on full images up to 2K resolution, ZipIR demonstrates a remarkable improvement over existing diffusion-based methods, highlighting the advantages of enhanced latent representation and scalable generative models for image restoration. We plan to explore even higher compression rates and larger capacity diffusion models for improved high-resolution image restoration.

{
    \small
    \bibliographystyle{ieeenat_fullname}
    \bibliography{main}
}

\clearpage

\appendix

\twocolumn[{%
    \centering
    \Large \textbf{Supplementary -- ``ZipIR: Latent Pyramid Diffusion Transformer for High-Efficiency High-Resolution Image Restoration"} \\
    \vspace{3em} 
}]

\section{Additional Implementation Details}
\noindent\textbf{Training Configuration.}
The optimization process employs AdamW with initial learning rate $5\times10^{-5}$ (decaying to $5\times10^{-6}$), weight decay 0.05, and betas $(0.9, 0.95)$. We implement DDPM loss with $\epsilon$-prediction objective, coupled with linear noise schedule ($\beta_\text{start}=0.00085$, $\beta_\text{end}=0.012$) and logit-normal time-step sampling for enhanced convergence.

\noindent\textbf{Architecture Specifications.}
Our LP-VAE is constructed as a UNet-based architecture with 128 base channels. Specifically, the encoders $E_c/E_f$ leverage residual blocks with channel multipliers $[1,2,4,4]/[1,2,4,4,4,4]$ respectively, and these configurations are mirrored in the decoders $D_c/D_f$. Each scale integrates two residual blocks powered by Swish activations. Our DiT adopts a 24-layer transformer, featuring a hidden dimension of 2048, 16 attention heads, and a strategy that incorporates adaptive layer normalization. Lastly, the f32c64-SDVAE used in ablation studies is modified from the LDM baseline, using $\text{z}_\text{channels}=64$ and $\text{ch}_\text{mult}=[1,2,4,4,4,4]$, thereby achieving a 32$\times$ spatial compression ratio.

\begin{figure}[b]
    \small
    \centering
    \includegraphics[width=0.9\linewidth]{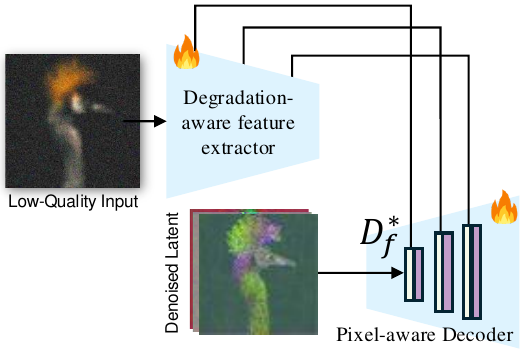}
        \vspace{-2mm}
    \caption{The pixel-aware decoder extracts high-res features from raw pixels with a degradation-aware feature extractor, enhancing the low-level fidelity of the decoded result during inference.
    }
    \label{fig:training-pipeline-pixelaware-decoder}
    % \vspace{-4mm}
\end{figure}

\noindent\textbf{Pixel-aware Decoder.}  
While previous work~\cite{wang2024stablesr} adopts a similar approach by fine-tuning auxiliary networks for skip connections within the VAE decoder, we propose a joint optimization of the feature extractor and VAE decoder. As illustrated in~\cref{fig:training-pipeline-pixelaware-decoder}, our pixel-aware decoder builds upon this design.

\section{Ultra-High Image Super-Resolution}

\begin{figure*}[t]
    \centering
    \includegraphics[page=1, width=\linewidth]{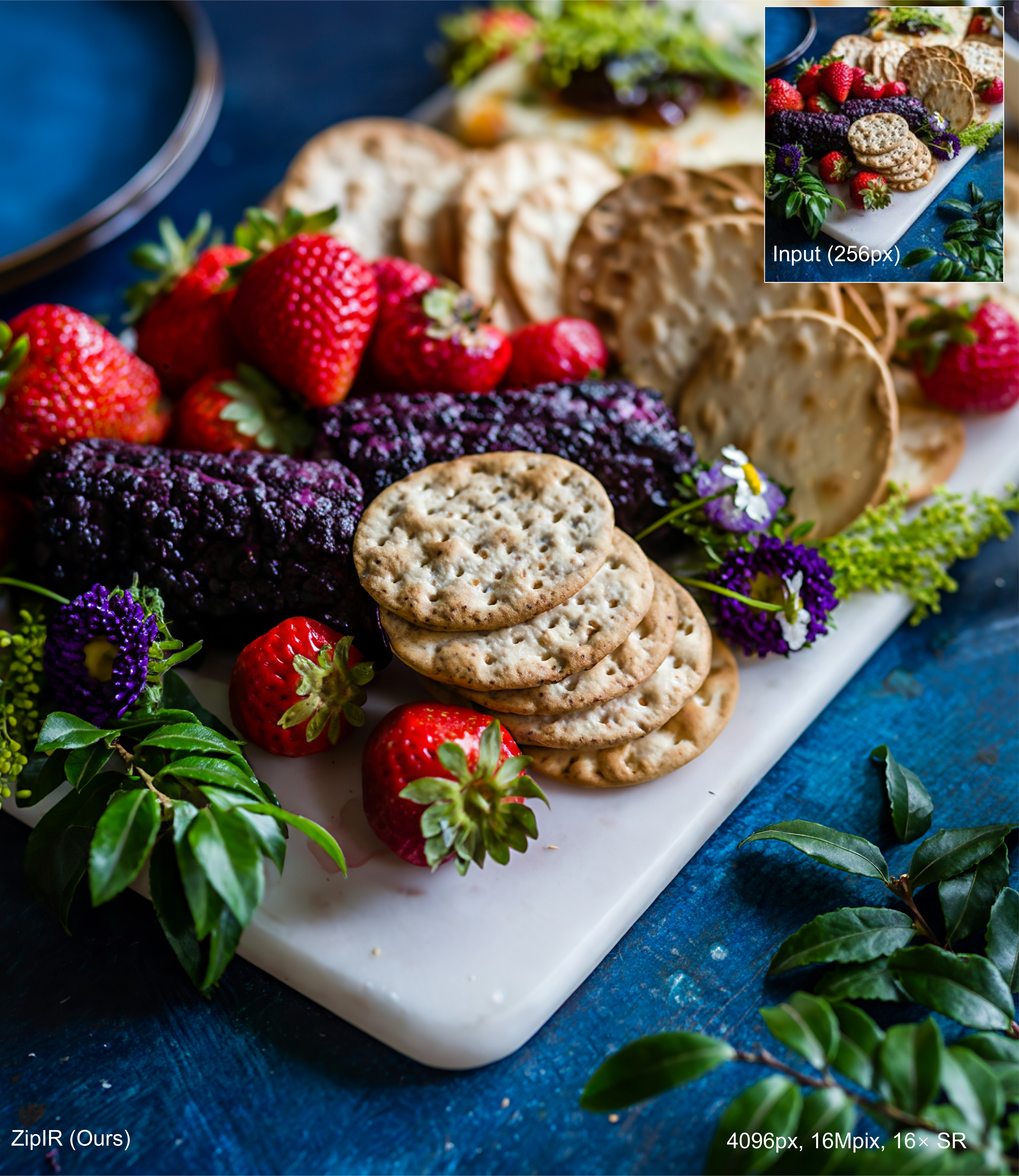}
    \caption{4K Image Super-Resolution Result by ZipIR. The input is a 256px low-resolution image, and the output achieves a 4096px (16Mpix) resolution with 16$\times$ scaling.}
    \label{fig:ultrahigh-cases-1}
\end{figure*}

\begin{figure*}[t]
    \centering
    \includegraphics[page=3, width=\linewidth]{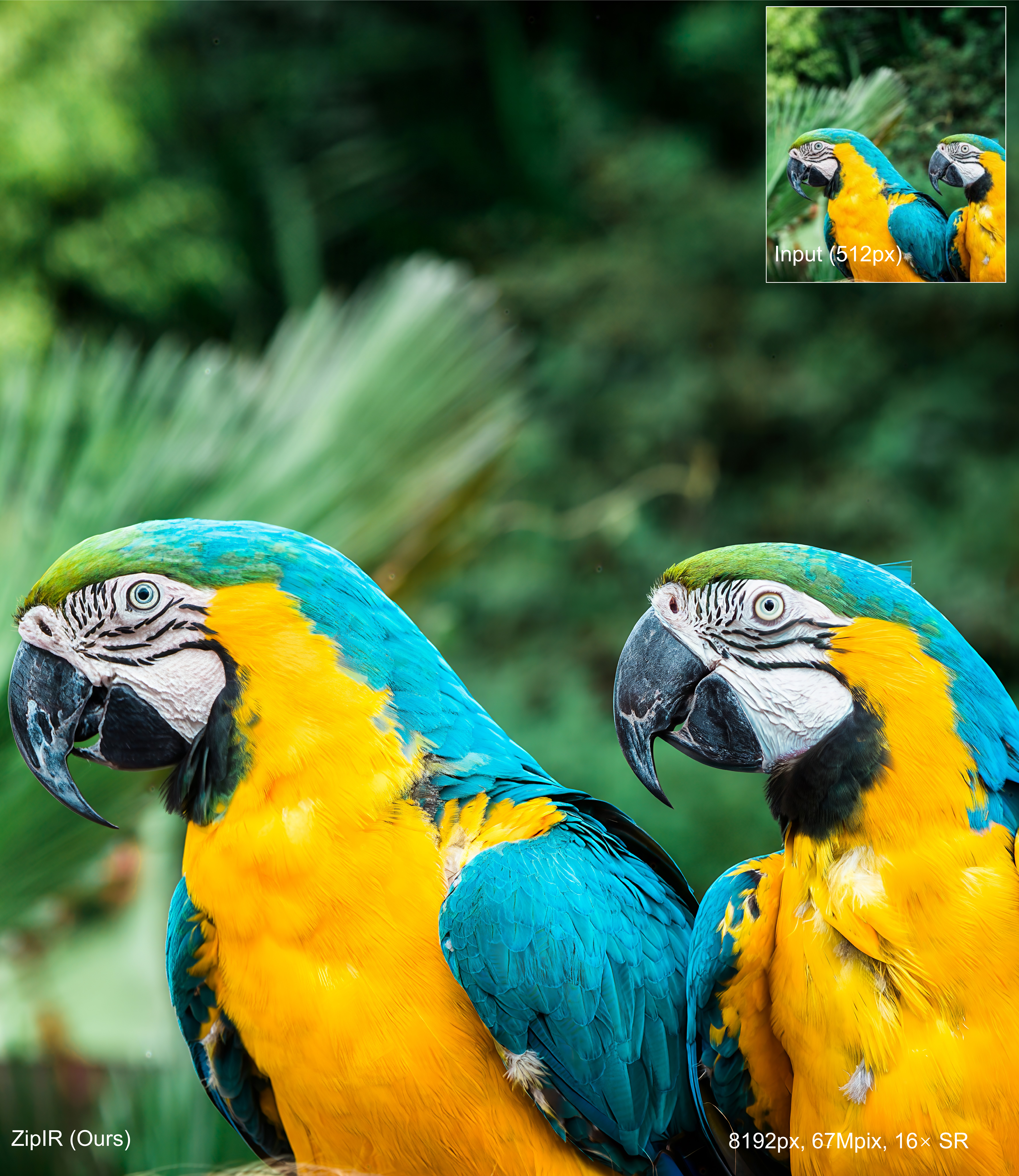}
    \caption{8K Image Super-Resolution Result by ZipIR. The input is a 512px low-resolution image, and the output achieves a 8192px (67Mpix) resolution with 16$\times$ scaling.}
    \label{fig:ultrahigh-cases-3}
\end{figure*}

In the main text, we comprehensively present cases of 2K image restoration. To further evaluate whether our method can generalize effectively to ultra-high-resolution image super-resolution, such as 4K and even 8K, we conducted additional experiments. As shown in Figures~\ref{fig:ultrahigh-cases-1}-\ref{fig:ultrahigh-cases-3}, our ZipIR achieved a 16$\times$ upscale, enhancing 256-pixel and 512-pixel images to 4K and 8K resolutions, respectively. This demonstrates the capability of our method to handle ultra-high-resolution image super-resolution effectively.

\begin{figure}[t]
    \centering
    \includegraphics[width=\linewidth]{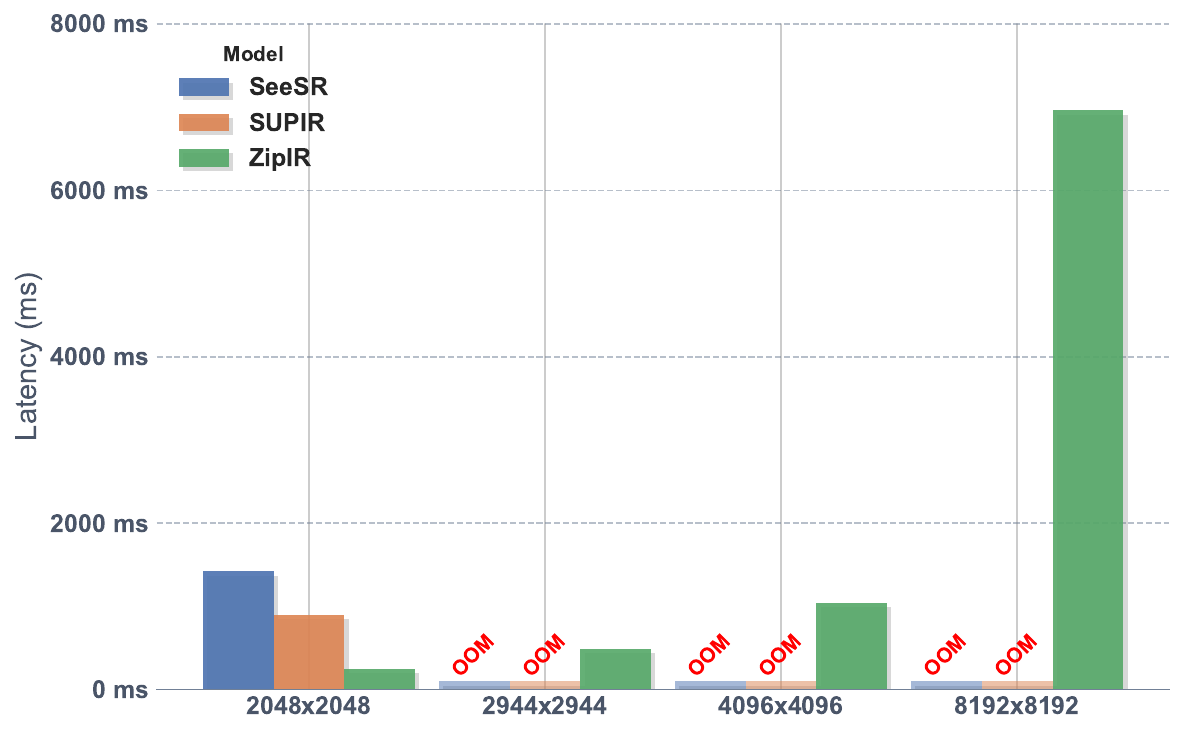}
    \caption{Denoising latency across ultra-high resolutions.}
    \label{fig:supp-ultra-high-latency}
\end{figure}

\begin{figure*}[t]
    \centering
    \begin{minipage}[b]{0.495\linewidth}
        \centering
        \includegraphics[page=1,width=\linewidth]{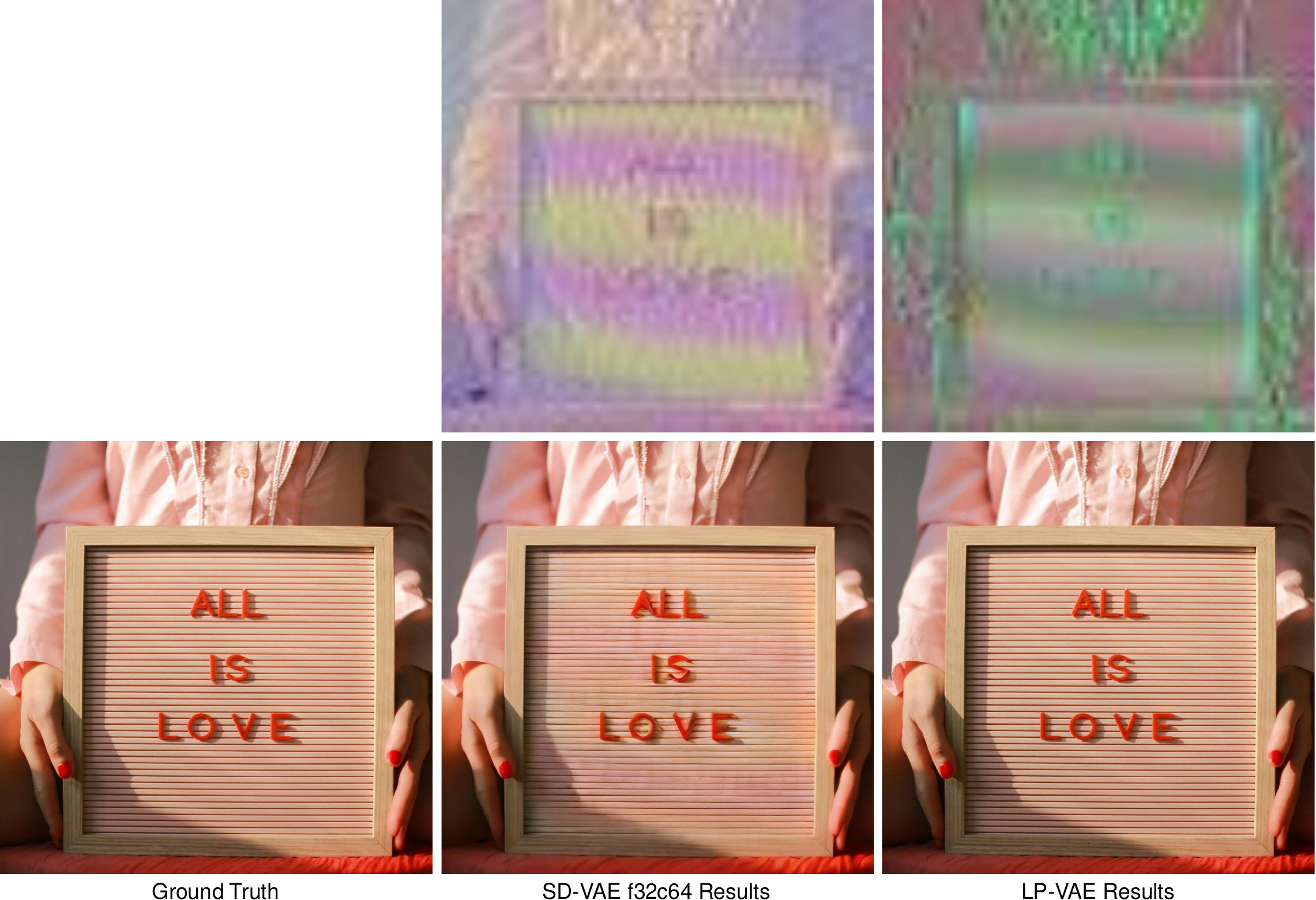}
    \end{minipage}
    % \hfill
    \begin{minipage}[b]{0.495\linewidth}
        \centering
        \includegraphics[page=2,width=\linewidth]{supp/comparison_vae_recon-crop.pdf}
    \end{minipage}
    \caption{Qualitative comparison of VAE image reconstruction. The first row visualizes the latent space representations, while the second row shows the reconstructed results.}
    \label{fig:supp_vae_recon_qualitative}
\end{figure*}

We have supplemented the evaluation of inference latency for ZipIR during diffusion denoising at each time step when synthesizing ultra-high-resolution images, including 4K and 8K resolutions. All efficiency evaluations were conducted on an A100-80G GPU. As shown in~\cref{fig:supp-ultra-high-latency}, our ZipIR demonstrates significant advantages across all resolutions. In contrast, the second-best method, SUPIR, fails to infer images larger than 2944$^2$ resolution due to out-of-memory errors. Furthermore, ZipIR achieves lower inference latency at 4096$^2$ resolution compared to SeeSR at 2048$^2$.

\section{Real-World HR Image Restoration}

\begin{figure*}[t]
    \centering
    \includegraphics[page=1, width=\linewidth]{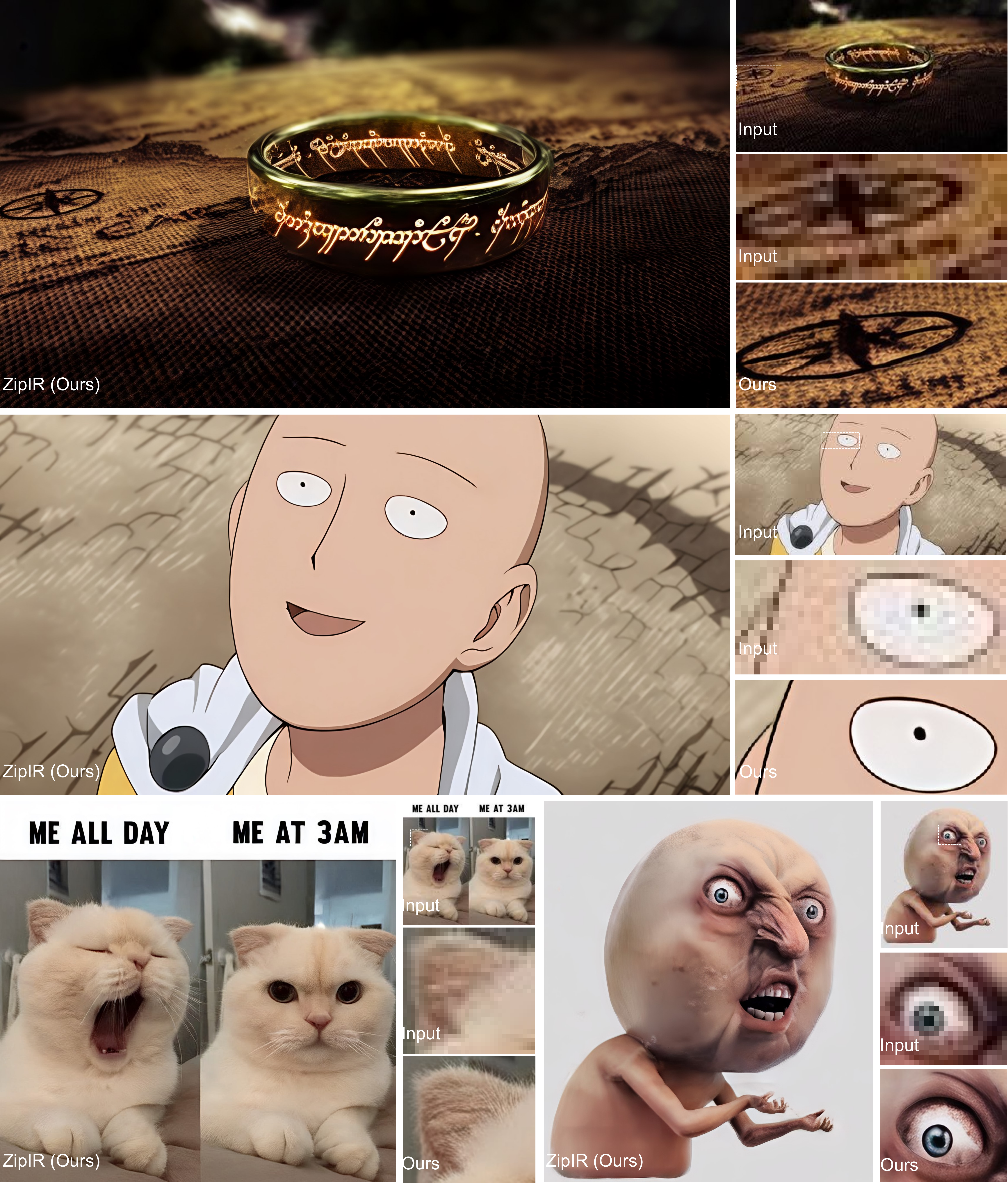}
    \caption{Real-world image restoration results by ZipIR. The inputs are low-resolution thumbnails sourced from the internet, featuring in-the-wild degradations.}
    \label{fig:real-world-cases-1}
\end{figure*}

\begin{figure*}[t]
    \centering
    \includegraphics[page=2, width=\linewidth]{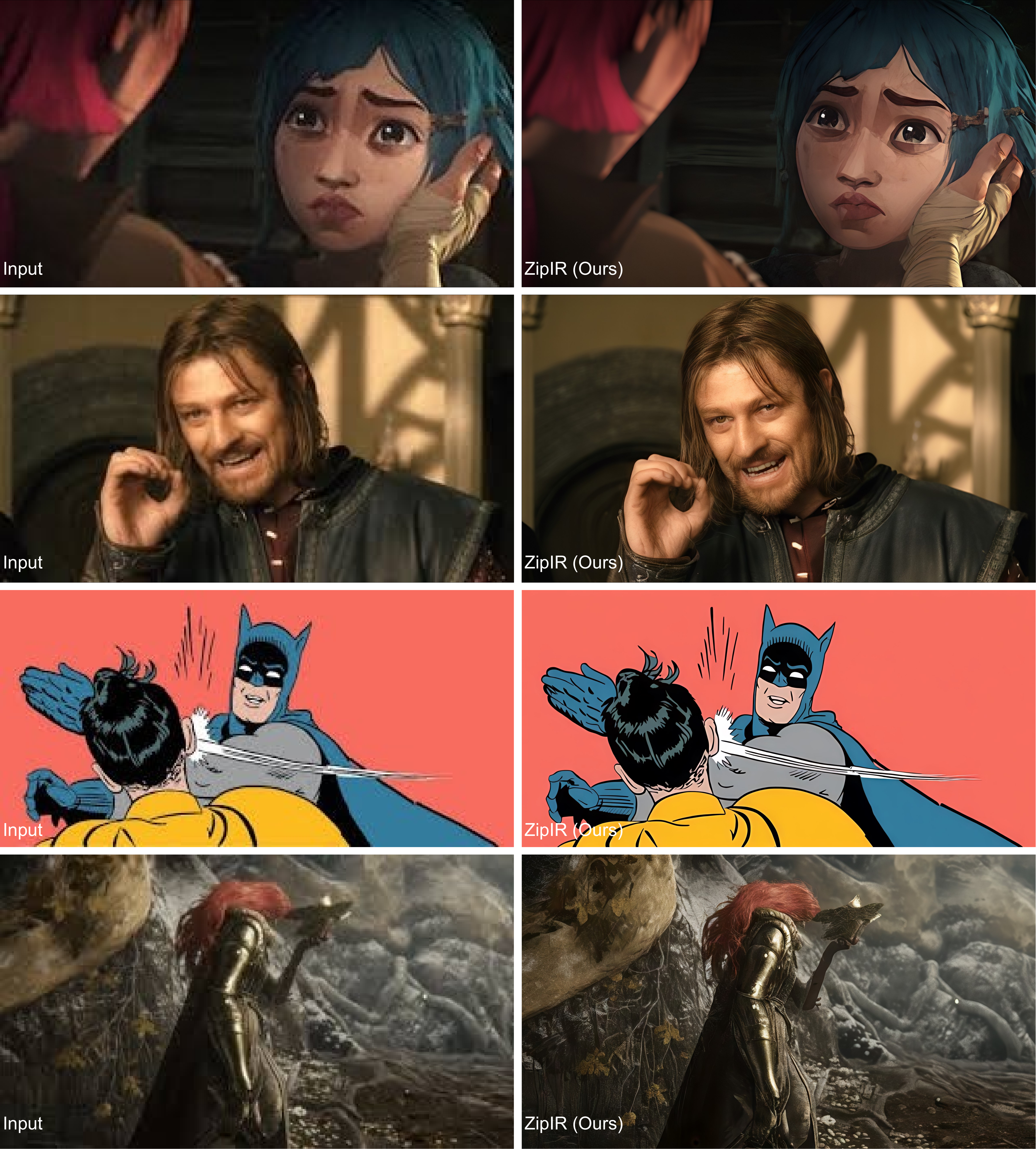}
    \caption{Real-world image restoration results by ZipIR. The inputs are low-resolution thumbnails sourced from the internet, featuring in-the-wild degradations.}
    \label{fig:real-world-cases-2}
\end{figure*}

We collected randomly sampled image thumbnails from the internet, capturing diverse real-world degradations, and used them as LQ inputs for high-resolution image restoration experiments. As shown in Figures~\ref{fig:real-world-cases-1} and~\ref{fig:real-world-cases-2}, our ZipIR effectively removes compression artifacts, reduces noise, and enhances fine details, producing high-resolution restored images with improved perceptual quality. These results demonstrate ZipIR’s robustness in handling in-the-wild degradations across varying content and degradation types, highlighting its practical applicability in real-world scenarios.

\section{Comparisons at a 4$\times$ Scale Factor}

\begin{table}[b]
% \small
\centering
\caption{Comparative analysis of 4$\times$ image restoration methods on real-world low-quality datasets.}
\label{tab:supp_4x_quantitative}
\resizebox{\linewidth}{!}{
\begin{tabular}{c|c|ccccc}
\toprule
Dataset & Method & FID$\downarrow$ & pFID$\downarrow$ & PSNR$\uparrow$ & SSIM$\uparrow$ & LPIPS$\downarrow$ \\
\midrule
\multirow{10}{*}{RealSR~\cite{cai2019toward}} 
& BSRGAN~\cite{zhang2021bsrgan}         & 78.17  & 95.84  & 25.35  & 0.7385 & 0.2809 \\
& Real-ESRGAN~\cite{wang2021realesrgan} & 82.41  & 88.18  & 24.70  & 0.7384 & 0.2865 \\
& SwinIR~\cite{liang2021swinir}         & 78.43  & 84.59  & 24.86  & \textbf{0.7444} & \textbf{0.2732} \\
& SD Upscaler~\cite{rombach2022high}    & 68.33  & 81.39  & 24.60  & 0.6644 & 0.3598 \\
& DiffBIR~\cite{lin2024diffbir}         & 68.75  & 90.62  & \textbf{25.62}  & 0.7149 & 0.3896 \\
& SeeSR~\cite{wu2024seesr}              & 68.96  & 79.42  & 25.06  & 0.7209 & 0.2874 \\
& SUPIR~\cite{yu2024supir}              & 61.84  & 84.99  & 24.04  & 0.6673 & 0.3425 \\
& Ours                                  & \textbf{61.35} & \textbf{78.07}  & 24.19  & 0.6999 & 0.2750 \\
\midrule
\multirow{10}{*}{DrealSR~\cite{wei2020drealsr}} 
& BSRGAN~\cite{zhang2021bsrgan}         & 41.57  & 69.96  & 24.88  & \textbf{0.6969} & \textbf{0.2174} \\
& Real-ESRGAN~\cite{wang2021realesrgan} & 45.10  & 71.51  & 24.05  & 0.6861 & 0.2273 \\
& SwinIR~\cite{liang2021swinir}         & 43.60  & 66.47  & 24.19  & 0.6905 & 0.2209 \\
& SD Upscaler~\cite{rombach2022high}    & 33.96  & 68.83  & 23.91  & 0.6276 & 0.2992 \\
& DiffBIR~\cite{lin2024diffbir}         & 36.33  & 62.19  & 25.03  & 0.6701 & 0.2175 \\
& SeeSR~\cite{wu2024seesr}              & 35.05  & 65.38  & 24.59  & 0.6701 & 0.2064 \\
& SUPIR~\cite{yu2024supir}              & 38.93  & 69.42  & 23.97  & 0.6193 & 0.2884 \\
& Ours                                  & \textbf{24.62} & \textbf{61.08} & \textbf{25.12} & 0.6843 & 0.2541 \\
\bottomrule
\end{tabular}}
\end{table}

Our goal is not to achieve state-of-the-art performance on traditional settings (e.g., medium-resolution outputs at a 4$\times$ scale factor on previous benchmarks), but rather to enable efficient image restoration (IR) at ultra-high resolutions. 

Nevertheless, for fair comparisons and a more comprehensive evaluation of our proposed method, we provide experiments on previous real-world image restoration benchmarks, including the DrealSR~\cite{wei2020drealsr} and RealSR~\cite{wei2020drealsr} datasets, for a 4$\times$ scale factor. Specifically, we restore LQ images of $256^2$ px to HQ resolutions of $1024^2$ px. Following the experimental setup described in the main text, we adopt FID, patch FID (pFID), PSNR, SSIM, and LPIPS as evaluation metrics.

As shown in Table~\ref{tab:supp_4x_quantitative}, even on traditional real-world image restoration benchmarks, our ZipIR achieves the best FID (61.35 and 24.62) and pFID (78.07 and 61.08) across both datasets, demonstrating superior perceptual quality. On DrealSR, it also achieves the highest PSNR (25.12), reflecting exceptional clarity and detail. These results validate the robustness and effectiveness of ZipIR for real-world 4$\times$ image restoration.

\section{VAE Reconstruction}

To intuitively highlight the differences between our proposed LP-VAE and a straightforward deepening of SD-VAE, we present a qualitative comparison of LP-VAE and SD-VAE f32c64 on 2048-resolution image reconstruction. Both LP-VAE and SD-VAE f32c64 perform 32$\times$ image compression and use a 64-channel dimensionality to represent the latent space. As shown in Figure~\ref{fig:supp_vae_recon_qualitative}, our proposed LP-VAE faithfully reconstructs the images, while SD-VAE f32c64 struggles to recover high-frequency details such as text and facial features.

\section{Additional Qualitative Comparisons}  

In the main text, we provide only two qualitative comparisons. To offer a more comprehensive and intuitive evaluation of our method's performance, we present additional qualitative results.  

\begin{figure*}
    \centering
    \includegraphics[page=2,width=\linewidth]{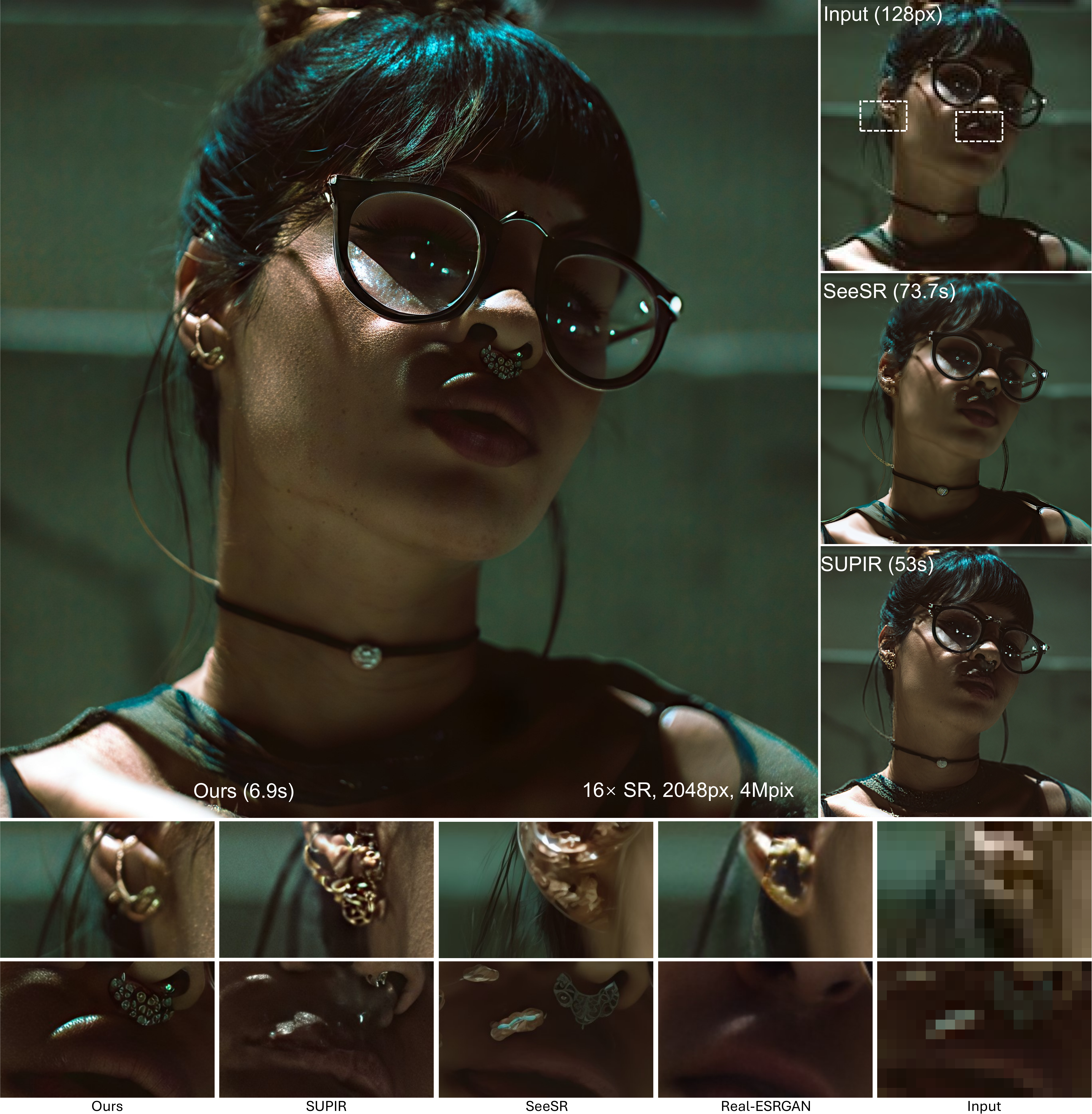}
    \caption{Our DiT-based ZipIR achieves 16$\times$ super-resolution, enhancing images from 128$^2$ to 2048$^2$ in just 6.9 seconds. Fine details, such as the nose ring and earrings, are faithfully restored without artifacts. Please zoom in for a detailed comparison.}
    \label{fig:addi_qualitative_1}
\end{figure*}

\begin{figure*}
    \centering
    \includegraphics[page=1,width=\linewidth]{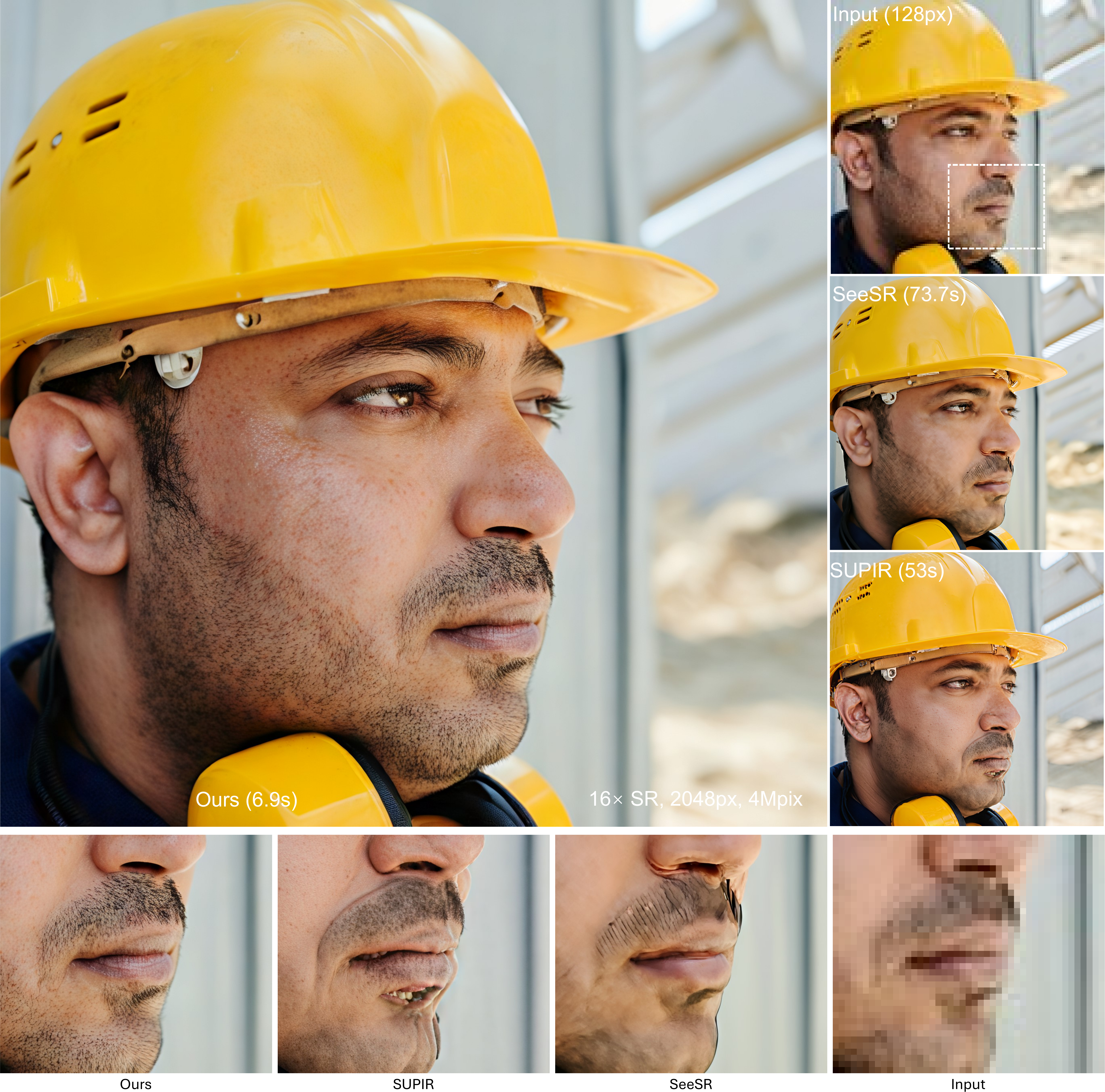}
    \caption{Our DiT-based ZipIR serves as a 16$\times$ upsampler, completing super-resolution from 128$^2$ to 2048$^2$ in only 6.9 seconds. Fine details, such as the chin, are accurately restored without artifacts. Please zoom in for a detailed comparison.}
    \label{fig:addi_qualitative_2}
\end{figure*}

\begin{figure*}
    \centering
    \includegraphics[page=3,width=\linewidth]{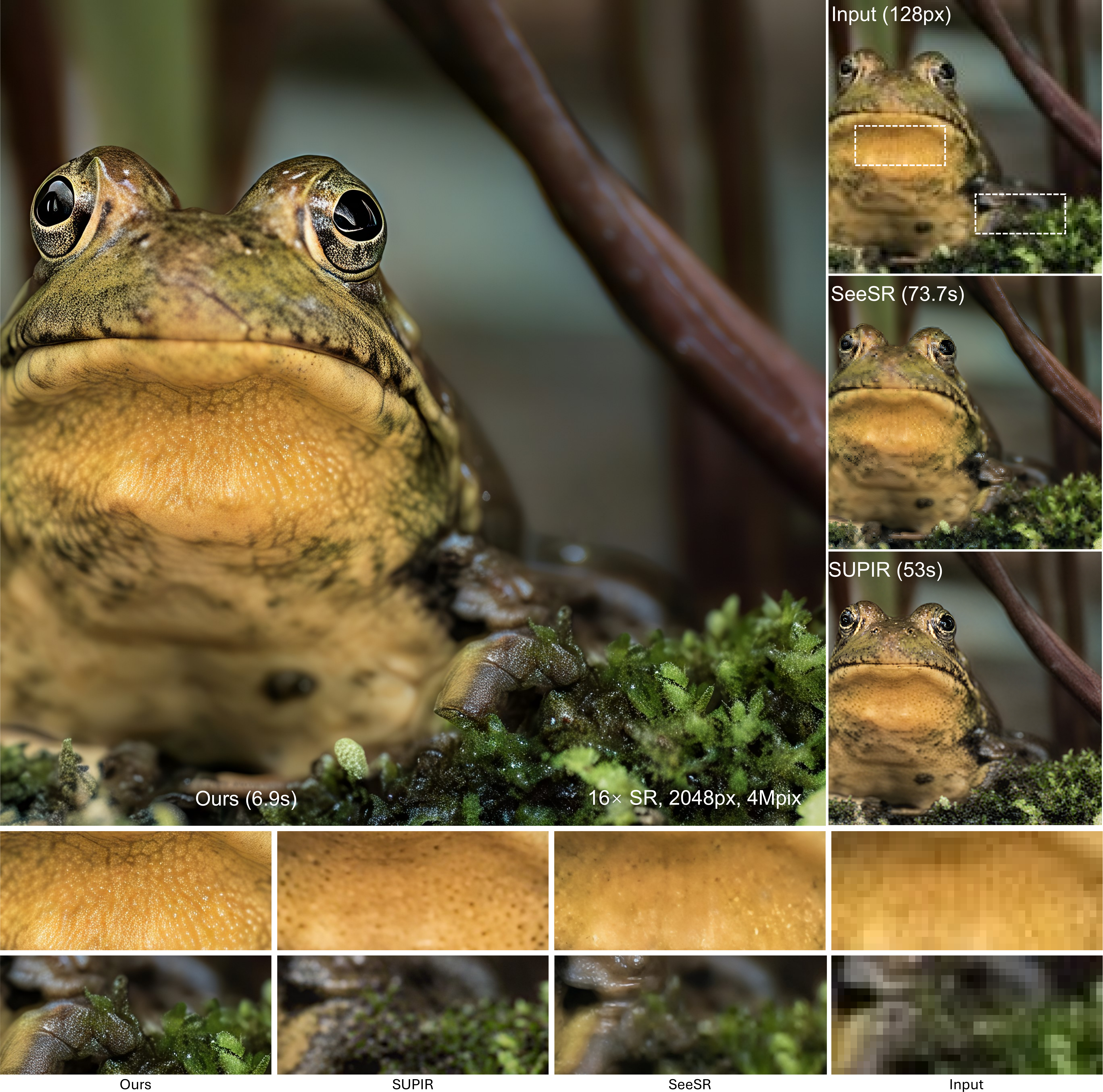}
    \caption{Our DiT-based ZipIR serves as a 16$\times$ upsampler, completing super-resolution from 128$^2$ to 2048$^2$ in only 6.9 seconds. The texture of the frog's chin is faithfully reconstructed without blur. Please zoom in for a detailed comparison.}
    \label{fig:addi_qualitative_3}
\end{figure*}

\noindent\textbf{2K 3000-Sample Test Set.}  
Following the experimental setup described in the main text, Figures~\ref{fig:addi_qualitative_1}-\ref{fig:addi_qualitative_3} showcase 16$\times$ super-resolution. Our ZipIR faithfully reconstructs fine details, such as the nose ring in Figure~\ref{fig:addi_qualitative_1}, and textures, like the frog's chin in Figure~\ref{fig:addi_qualitative_3}, while avoiding hallucination of unrealistic structures, as seen in the human chin in Figure~\ref{fig:addi_qualitative_2}.  

\begin{figure*}
    \centering
    \includegraphics[page=4,width=\linewidth]{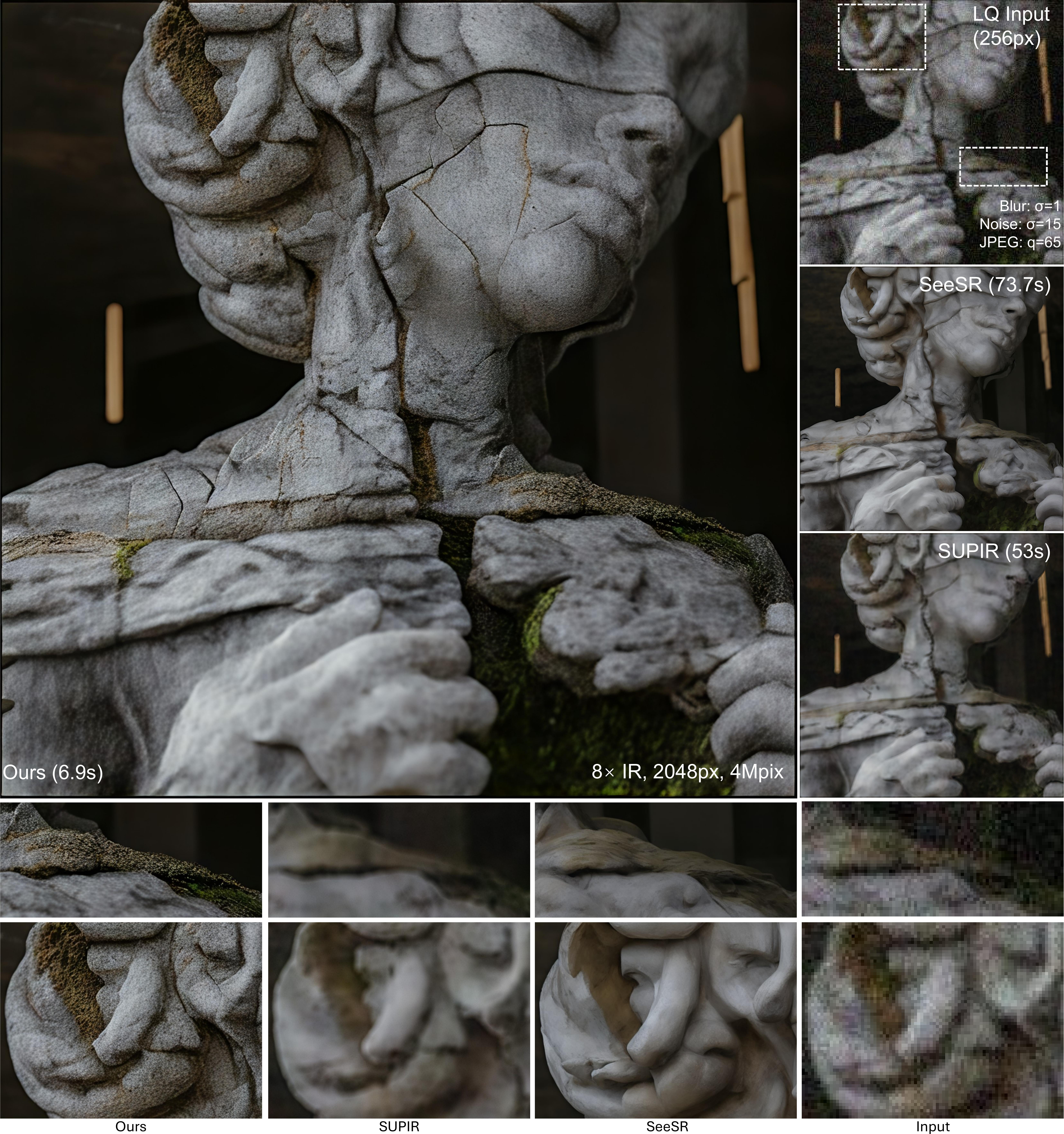}
    \caption{Our DiT-based ZipIR performs 8$\times$ super-resolution, enhancing images from 256$^2$ to 2048$^2$ in just 6.9 seconds, while simultaneously restoring details through deblurring, denoising, and JPEG artifact removal. The input image is degraded with Gaussian blur ($\sigma$ = 1), noise ($\sigma$ = 15), and JPEG compression (q = 65). Fine textures, such as the grass, are faithfully reconstructed. Zoom in for detailed comparison.}
    \label{fig:addi_qualitative_4}
\end{figure*}

\begin{figure*}
    \centering
    \includegraphics[page=5,width=\linewidth]{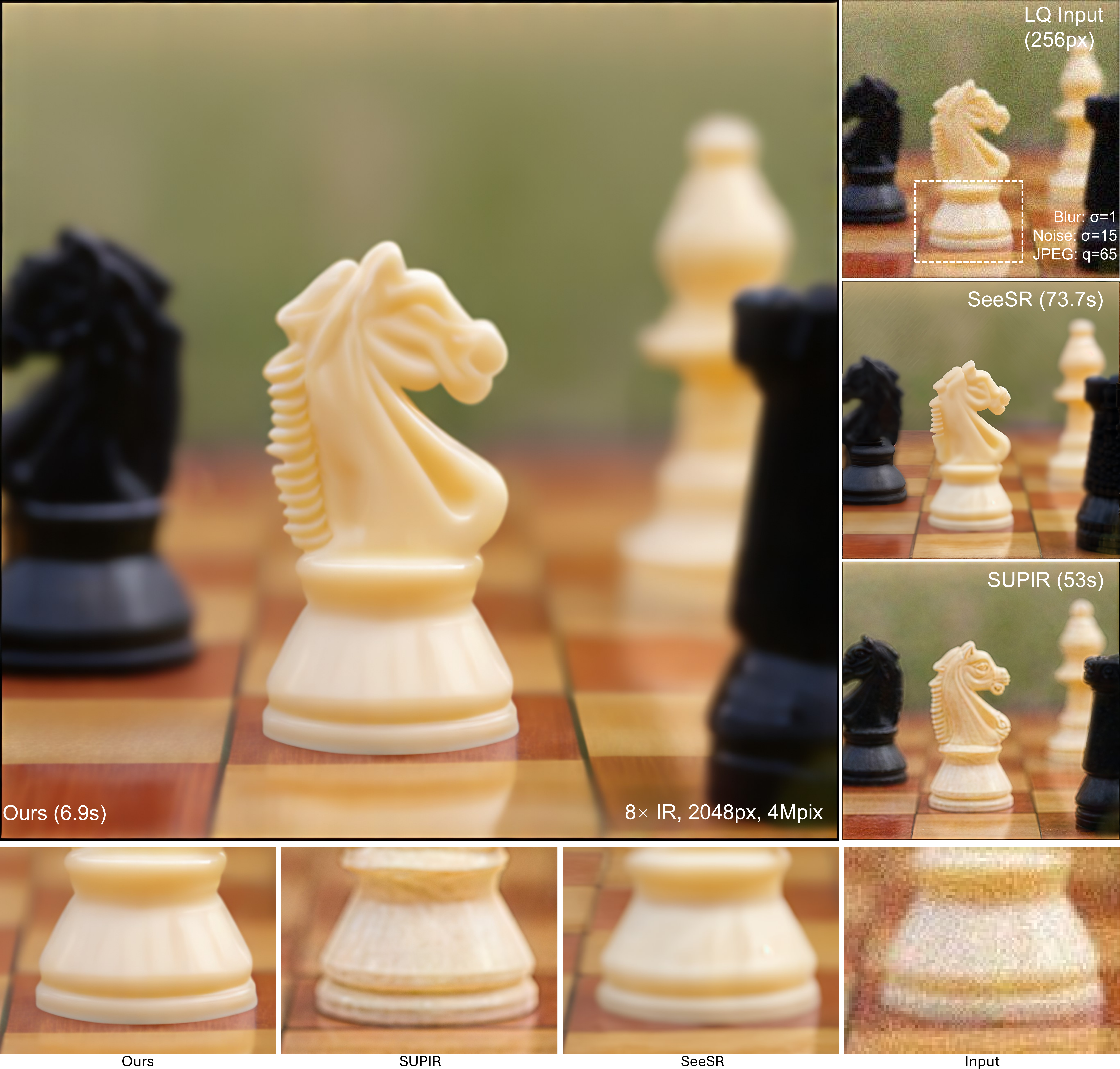}
    \caption{Our DiT-based ZipIR performs 8$\times$ super-resolution, enhancing images from 256$^2$ to 2048$^2$ in just 6.9 seconds, while simultaneously restoring details through deblurring, denoising, and JPEG artifact removal. The input image is degraded with Gaussian blur ($\sigma = 1$), noise ($\sigma = 15$), and JPEG compression ($q = 65$). The generated chess piece base is sharp and free from blurring.}
    \label{fig:addi_qualitative_5}
\end{figure*}

Figures~\ref{fig:addi_qualitative_4}-\ref{fig:addi_qualitative_5} illustrate 8$\times$ image restoration, where the inputs suffer from blur ($\sigma$=1), noise ($\sigma$=15), and JPEG artifacts (q=65). ZipIR effectively recovers realistic textures, such as the grass in Figure~\ref{fig:addi_qualitative_4}, and restores sharper details, exemplified by the chess piece in Figure~\ref{fig:addi_qualitative_5}.  

\begin{figure*}
    \centering
    \includegraphics[width=\linewidth]{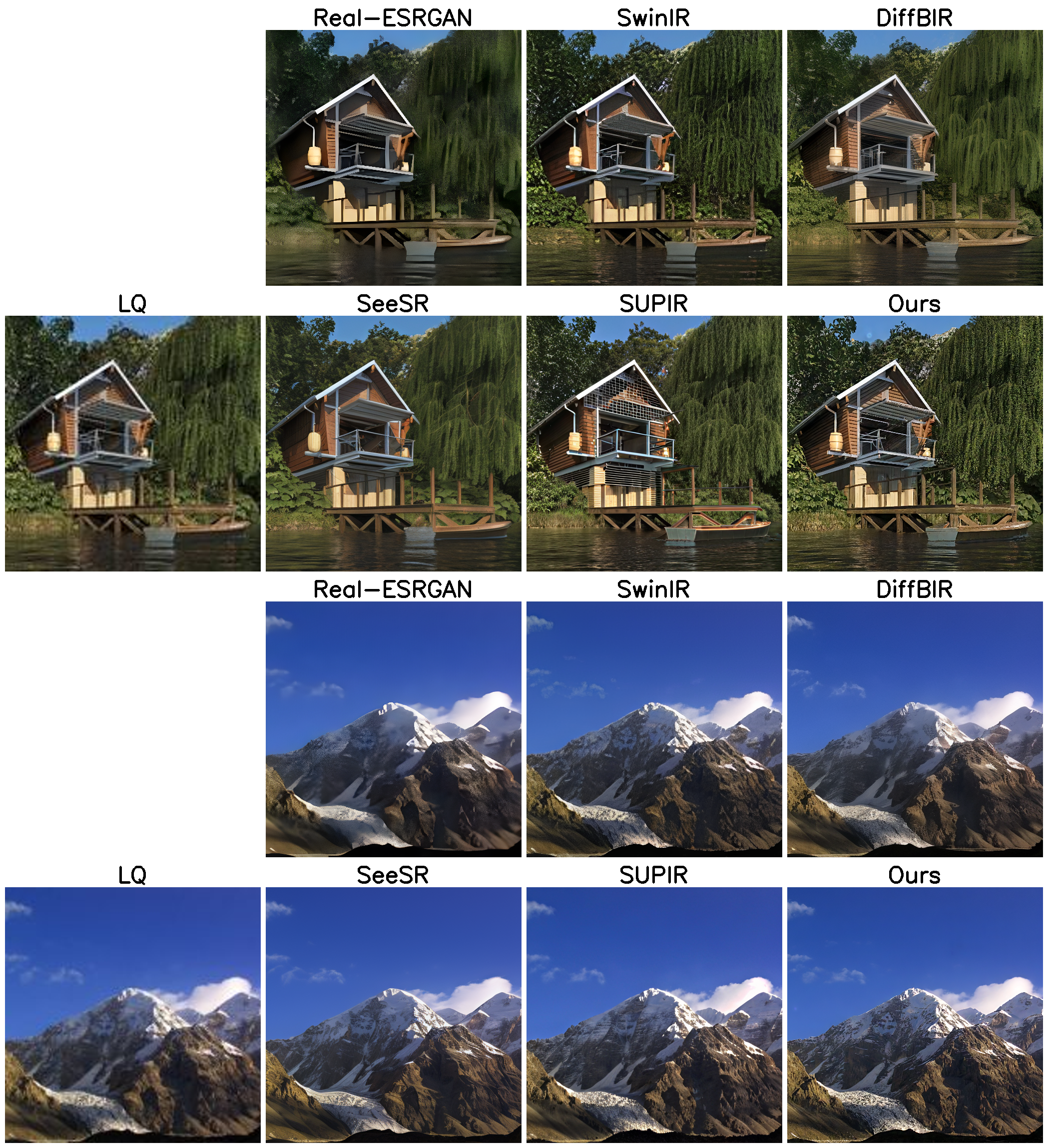}
    \caption{The visual comparison on the real-world LQ dataset RealPhoto60 dataset~\cite{yu2024supir} at resolution 512$^2$.}
    \label{fig:supp_realphoto_1}
\end{figure*}

\begin{figure*}
    \centering
    \includegraphics[width=\linewidth]{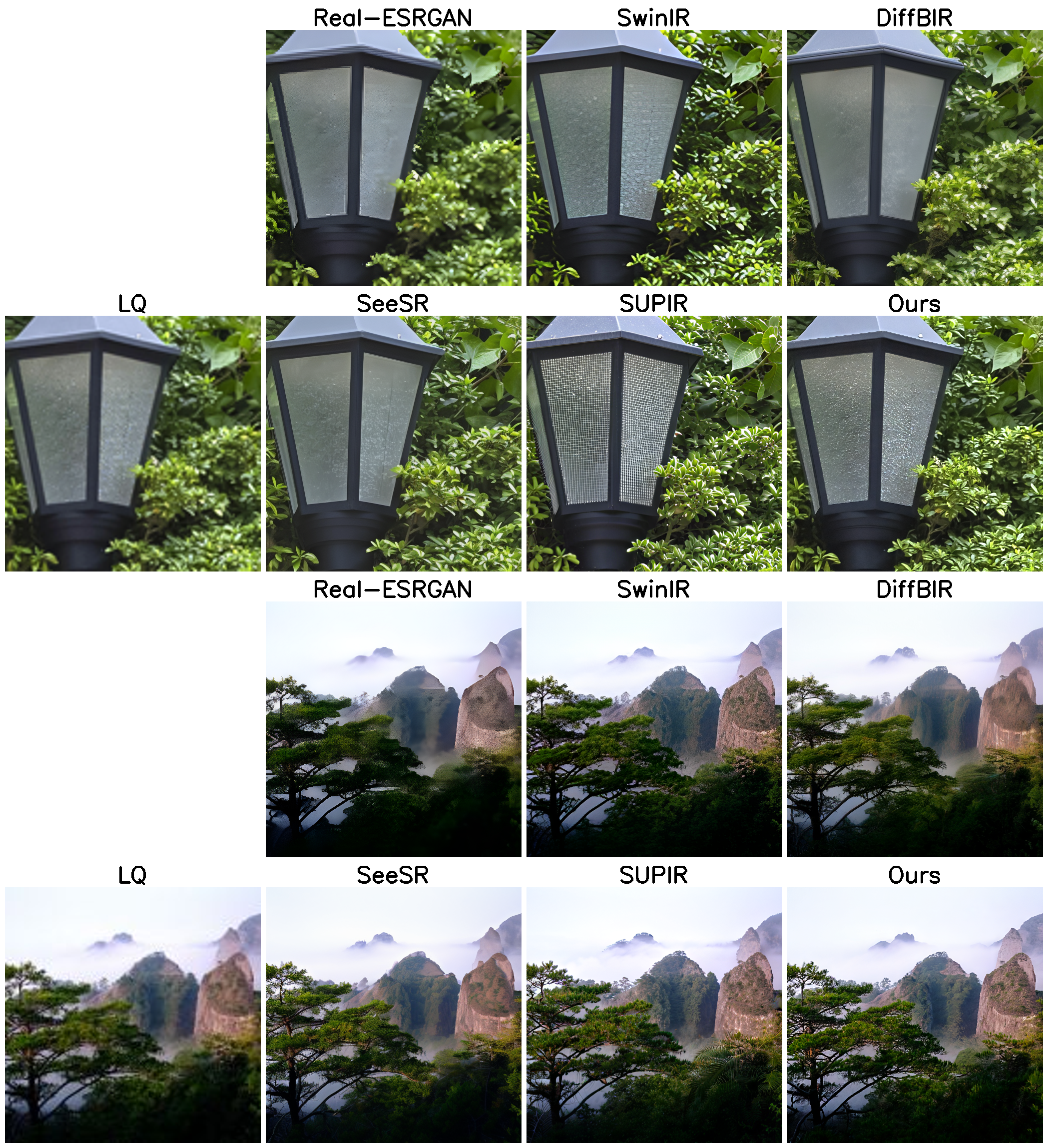}
    \caption{The visual comparison on the real-world LQ dataset RealPhoto60 dataset~\cite{yu2024supir} at resolution 512$^2$.}
    \label{fig:supp_realphoto_2}
\end{figure*}

\begin{figure*}
    \centering
    \includegraphics[width=\linewidth]{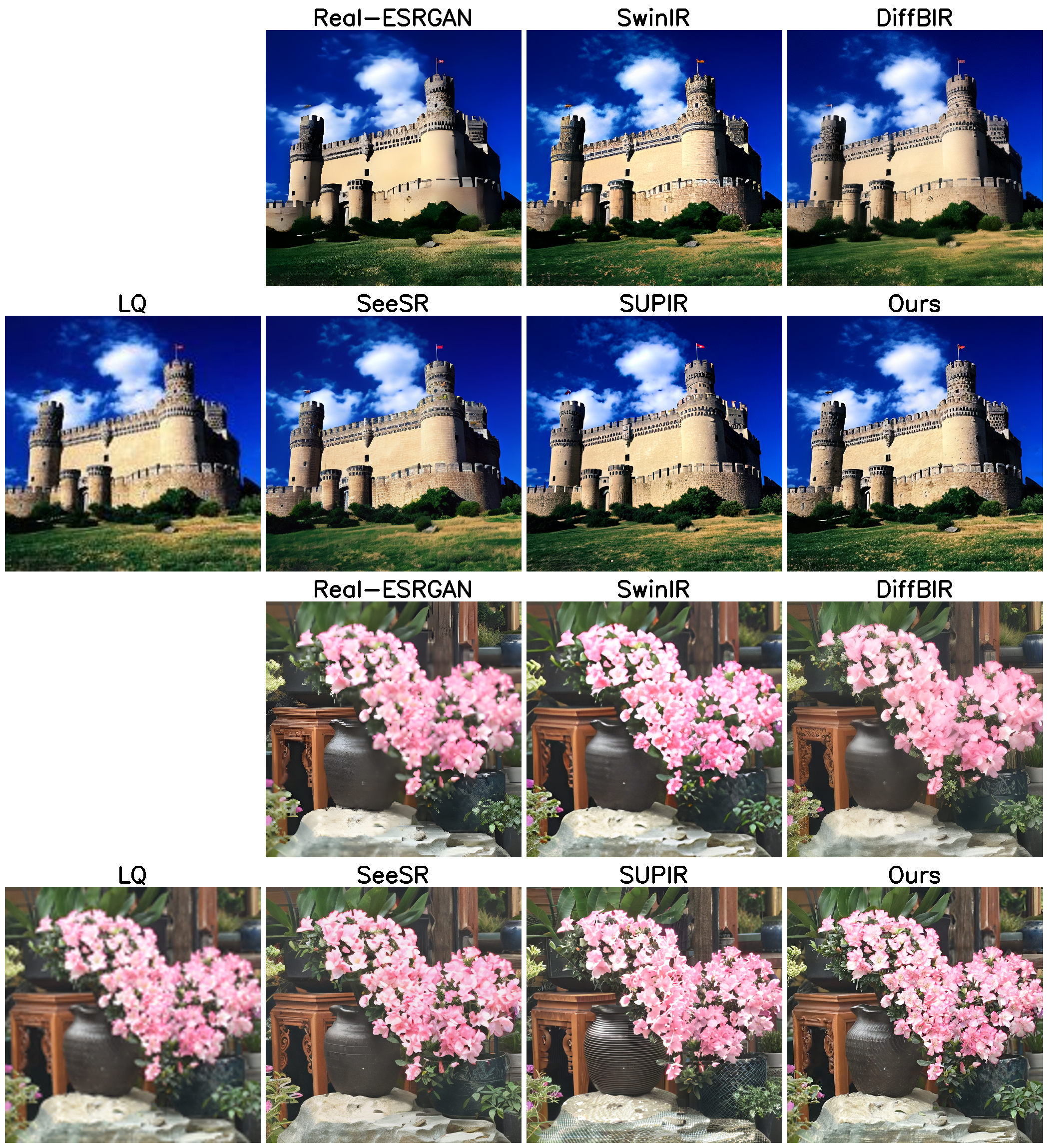}
    \caption{The visual comparison on the real-world LQ dataset RealPhoto60 dataset~\cite{yu2024supir} at resolution 512$^2$.}
    \label{fig:supp_realphoto_3}
\end{figure*}

\noindent\textbf{512px RealPhoto60~\cite{yu2024supir}.}  
As a qualitative counterpart to Table~2 in the main text, Figures~\ref{fig:supp_realphoto_1} to~\ref{fig:supp_realphoto_3} present visual comparisons on the RealPhoto60 test set. SeeSR and DiffBIR use their default training resolutions, while both SUPIR and our ZipIR process images at 1K resolution, following SUPIR’s approach, before downsampling to 512px.

\section{Effectiveness of Text Prompt}

Table~\ref{tab:supp_text_prompt_ablation} compares configurations with and without text prompts, highlighting the influence of CFG strength on FID, pFID, and PSNR. The default setting, with a CFG strength of 3.5, achieves a balanced performance, yielding an FID of 18.05, pFID of 34.85, and PSNR of 27.75. Reducing CFG strength to 1.5 slightly enhances PSNR but worsens FID and pFID, while increasing it to 5.5 maximizes PSNR (28.25) at the expense of perceptual fidelity.

Without text prompts, FID and pFID degrade, though PSNR (28.04) marginally surpasses the default. This underscores the role of text prompts in enhancing perceptual fidelity, while CFG strength tuning mediates the trade-off between fidelity and quality.

\section{Test Set Construction}
\begin{figure}
    \centering
    \includegraphics[width=\linewidth]{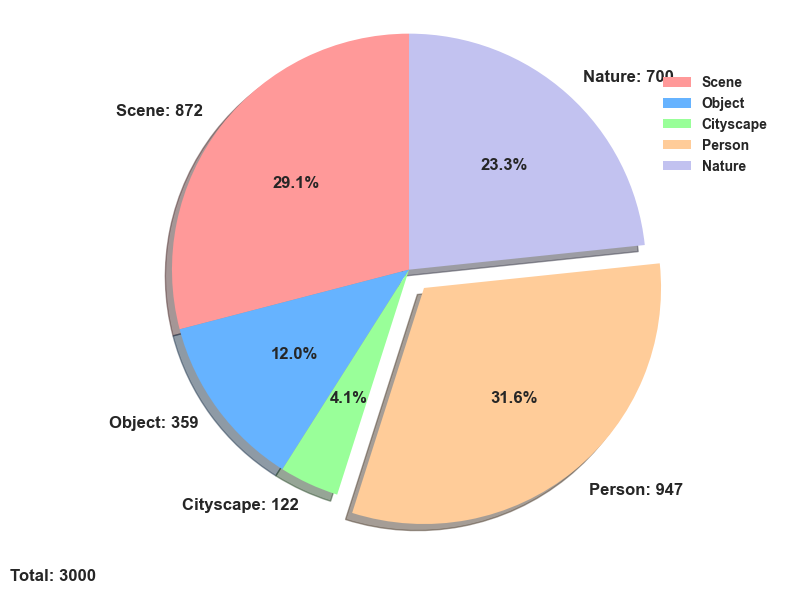}
    \caption{Semantic category distribution of our 3,000-sample test dataset, classified using CLIP~\cite{radford2021learning}.}
    \label{fig:category-test-set}
\end{figure}

We curate a high-resolution test set of 3,000 images randomly sampled from Pexels, ensuring comprehensive coverage of real-world visual concepts. The dataset encompasses diverse semantic content, validated through CLIP~\cite{radford2021learning} text similarity analysis, as illustrated in~\cref{fig:category-test-set}, spanning six major categories. Special attention is given to maintaining a balanced representation across visual domains while preserving natural image statistics.

\begin{table}
\centering
\small
\caption{Quantitative evaluation of configurations with and without text prompts, including varying CFG strengths under the `w/ text prompt` setting. The default configuration is `CFG strength = 3.5`, and its values are equivalent to those for `w/ text prompt`.}
\label{tab:supp_text_prompt_ablation}
\resizebox{\linewidth}{!}{%
\begin{tabular}{llccc}
\toprule
\multirow{2}{*}{Text Prompt} & \multirow{2}{*}{CFG Strength} & \multicolumn{3}{c}{Metrics} \\ \cmidrule(l){3-5} 
                             &                              & FID $\downarrow$ & pFID $\downarrow$ & PSNR $\uparrow$ \\ \midrule
\multirow{3}{*}{\textbf{w/ text prompt}} 
                             & Default (3.5)                & \textbf{18.05} & \textbf{34.85} & 27.75 \\
                             & 1.5                          & 19.73 & 35.12 & 27.90 \\
                             & 6.5                          & 20.41 & 36.88 & \textbf{28.25} \\ \midrule
\textbf{w/o text prompt}     & --                           & 19.97 & 35.40 & 28.04 \\ \bottomrule
\end{tabular}%
}
\end{table}

\clearpage

\end{document}